\pdfoutput=1

\documentclass[11pt]{article}

\usepackage[final]{EACL2023}

\usepackage[whole]{bxcjkjatype}

\usepackage{times}
\usepackage{latexsym}

\usepackage{url}
\usepackage{booktabs}
\usepackage{graphicx,xcolor}
\usepackage{amsmath}
\usepackage{amssymb}
\usepackage{ascmac}

\usepackage[T1]{fontenc}

\usepackage[utf8]{inputenc}

\usepackage{microtype}

\usepackage{inconsolata}

\usepackage{multirow}

\usepackage{colortbl}

\usepackage[framemethod=TikZ]{mdframed}

\newcommand{\graybox}[1]{\begin{mdframed}[
    backgroundcolor=black!5,
    topline=false, bottomline=false, rightline=false, leftline=false,
    innertopmargin=0.5em,
    innerleftmargin=0.5em,
    innerbottommargin=0.5em,
    innerrightmargin=0.5em,
]{#1}
\end{mdframed}
}

\title{Do Deep Neural Networks Capture Compositionality \\ in Arithmetic Reasoning?}

\author{Keito Kudo${}^{1}$
        Yoichi Aoki${}^{1}$ 
        Tatsuki Kuribayashi${}^{1,2}$
        Ana Brassard${}^{3,1}$ \\ 
        {\bf Masashi Yoshikawa${}^{1}$ Keisuke Sakaguchi${}^{1,3}$ Kentaro Inui${}^{1,3}$} \\
         ${}^{1}$Tohoku University
         ${}^{2}$Langsmith, Inc.
         ${}^{3}$RIKEN  \\ 
        \texttt{\{keito.kudo.q4, youichi.aoki.p2\}@dc.tohoku.ac.jp, }
        \texttt{kuribayashi@tohoku.ac.jp, } \\
        \texttt{ana.brassard@riken.jp, }
        \texttt{\{yoshikawa,keisuke.sakaguchi,inui\}@tohoku.ac.jp}}

\begin{document}
\maketitle
\begin{abstract}
Compositionality is a pivotal property of symbolic reasoning.
However, how well recent neural models capture compositionality remains underexplored in the symbolic reasoning tasks.
This study empirically addresses this question by systematically examining recently published pre-trained seq2seq models with a carefully controlled dataset of multi-hop arithmetic symbolic reasoning. 
We introduce a \textit{skill tree} on compositionality in arithmetic symbolic reasoning that defines the hierarchical levels of complexity along with three compositionality dimensions: systematicity, productivity, and substitutivity. 
Our experiments revealed that among the three types of composition, the models struggled most with systematicity, performing poorly even with relatively simple compositions.
That difficulty was not resolved even after training the models with intermediate reasoning steps.\footnote{Our code and data are available at \url{https://github.com/keitokudo/dentaku_skill_tree}.}

\end{abstract}

\section{Introduction}
Integrating symbolic reasoning capabilities into neural models has been a crucial goal of artificial intelligence~\cite{marcus2003algebraic,DBLP:journals/corr/abs-2012-05876}.
With this in mind, many researchers investigated how well modern neural models achieve symbolic reasoning~\cite{Lake2018GeneralizationWS}.
However, recent studies have reported conflicting results on this; some suggest that neural models can solve complex multi-hop reasoning~\cite{ijcai2020p537}, while others claim that models struggle even with performing simple symbolic operations~\cite{DBLP:journals/corr/abs-2208-05051}.

\begin{figure}[t]
\centering
\includegraphics[width=\linewidth]{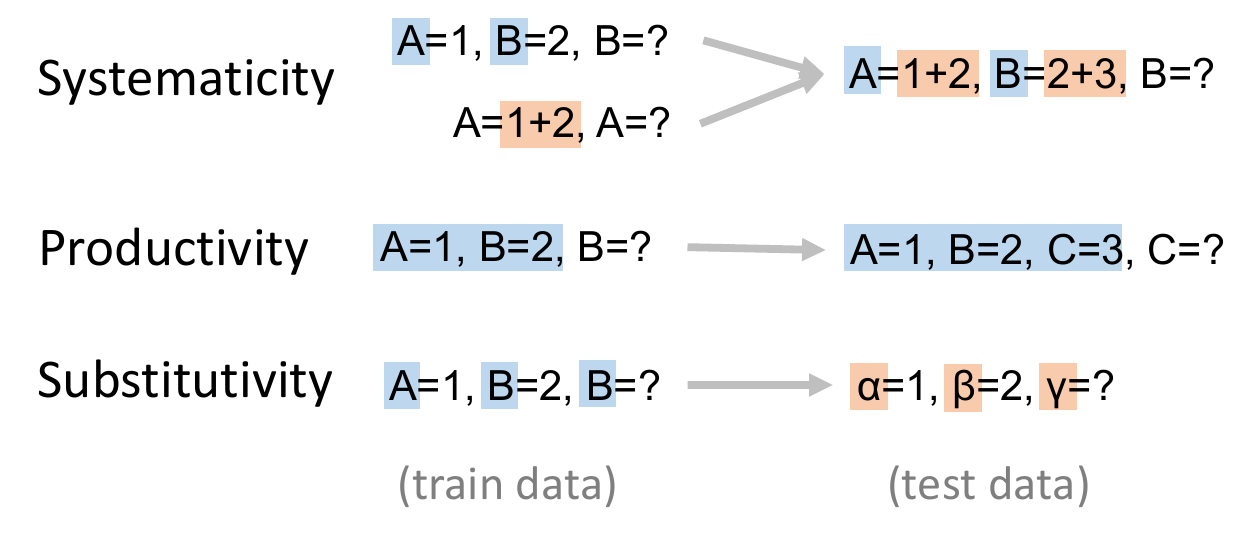}
\caption{Three dimensionalities of compositionality in arithmetic symbolic reasoning}
\label{fig:overview}  
\end{figure}

As a step toward further understanding neural models' symbolic reasoning ability, this study systematically analyzes recently published pre-trained seq2seq models using a carefully controlled dataset of multi-hop arithmetic symbolic reasoning. 
Specifically, our study empirically evaluates the models' ability to generalize the \emph{compositionality} underlying arithmetic reasoning, where we explore three dimensions of compositionality: (i) systematicity, (ii) productivity, and (iii) substitutivity,  as illustrated in Figure~\ref{fig:overview}.
Capturing compositionality is crucial in performing symbolic reasoning since compositionality is a pivotal property of generalizability over training instances.

To systematically explore the models' composition ability, we introduce a \textit{skill tree on compositionality} that defined the hierarchical levels of complexity in arithmetic symbolic reasoning, as illustrated in Figure~\ref{fig:skill-tree}.
Using this hierarchy as a lens, we identify the limitations of the neural seq2seq models in capturing the compositionality in arithmetic symbolic reasoning.
Our major findings can be summarized as follows:

\vspace{-0.5\baselineskip} 
\begin{itemize}
  \setlength{\parskip}{0cm} %
  \setlength{\itemsep}{0cm} %
  \item Among the three types of composition, the models struggled most with \textbf{systematicity}, performing poorly even with relatively simple compositions.
  \item The major difficulty in systematicity was in the access to intermediate information that is not stated in input but produced during the reasoning.
  \item Capturing systematicity remained hard for the models trained with the information of the intermediate reasoning steps. 
\end{itemize}
\vspace{-0.5\baselineskip} 

\section{Skill tree in arithmetic reasoning}
\label{sec:problem}
We take arithmetic reasoning as the domain for our exploration because it allows us to synthesize questions systematically, as we show in this paper, which helps examine a model's composition ability in a controlled manner. 
Furthermore, the arithmetic reasoning ability of neural models has gained much attention as modern large language models still struggle with this problems~\cite {DBLP:journals/corr/abs-2112-11446}.

Specifically, we use multi-hop arithmetic reasoning problems as follows:
\graybox{
 \emph{Question:} \texttt{A=1, B=2, C=A+2, C=?} \\
 \emph{Answer:} \texttt{3}
}

\noindent
Here, the value assigned to the variable \texttt{C} is asked.

\subsection{Compositionality in multi-hop symbolic reasoning}
\label{subsec:compositionality_in_multi-hop_symbolic_reasoning}
In this study, we specifically focused on three 
dimensions: systematicity, productivity, and substitutivity~\cite{ijcai2020p708}.
According to these, we evaluate how well neural models achieve compositional generalization.

\noindent
\textbf{Systematicity}
\label{subsec:systematicity_generalization}
refers to the ability of combining known different concepts into a more complex concept, i.e., structural composition.
To evaluate this ability in models, we first trained with several types of primitive operations (e.g., addition; \texttt{A=1+2,A=?} and selection; \texttt{A=1,B=2,B=?}).
Then, we measured the performance in solving problems consisting of combinations of primitives (e.g., \texttt{A=1+2,B=2+3,B=?}).

\noindent
\textbf{Productivity}
\label{subsec:productivity_generalization}
 refers to the ability to solve longer/complex problems based on shorter/simpler ones.
To evaluate this ability in models, we first trained with a short version of a formula (e.g., \texttt{A=1+2,B=2+3,B=?}).
Then, we measured the performance in solving longer problems (e.g., \texttt{A=1+2,B=2+3,C=3+4,C=?}).

\noindent
\textbf{Substitutivity}
\label{subsec:substitutivity_generalization}
 refers to the ability to keep the performance even if a particular constituent in a problem is replaced with another (unseen) constituent (i.e., lexical composition).
To evaluate this ability in models, we conduct several experiments changing the variable characters between training and test (e.g., train with \texttt{A=1+2,A=?}; then evaluated with \texttt{α=1+2,α=?}).

\subsection{Dataset configurations}
\label{subsec:dataset_configurations_and_evaluation_methods}
Typical symbolic reasoning (e.g., procedural programming, assembly language) consists of at least three primitive symbol manipulations: assignment (\texttt{a=2}), arithmetic operation (\texttt{1+2}), and reference (\texttt{a=?}).
With this in mind, our dataset is generated by combining the following five basic formulas: (i) \texttt{A=1} (assignment), (ii) \texttt{A=B} (reference \& assignment), (iii) \texttt{A=1+2} (arithmetic operation \& assignment), (iv) \texttt{A=B+2} (arithmetic operation \& assignment \& reference), (v) \texttt{A=?} (reference).
The detailed properties are explained in Section~\ref{sec:evaluation_method}.

\subsection{Skill tree evaluations}
\label{subsec:skill-tree-evaluation}
We preliminarily observed that compositionally generalizing complex multi-hop arithmetic reasoning was difficult for neural seq2seq learners (the 1,2,6$\rightarrow$9 setting in Section~\ref{section:results}).
Building on this fact, this study questions what type of composition made it hard for the neural models.
To answer this, we designed a \textit{skill tree on compositionality} that organizes the (hierarchical) complexity levels of symbolic reasoning.\footnote{The term "skill tree" refers to a visualization method of step-by-step learning in the field of pedagogy~\cite{22136}, distinct from the "tree" in graph theory.}
Evaluating the models using problems with different complexity of composition in a step-wise manner, we elucidate the exact weakness of neural seq2seq models in multi-hop symbolic reasoning.

\begin{figure}[!t]
\centering
\includegraphics[width=\linewidth]{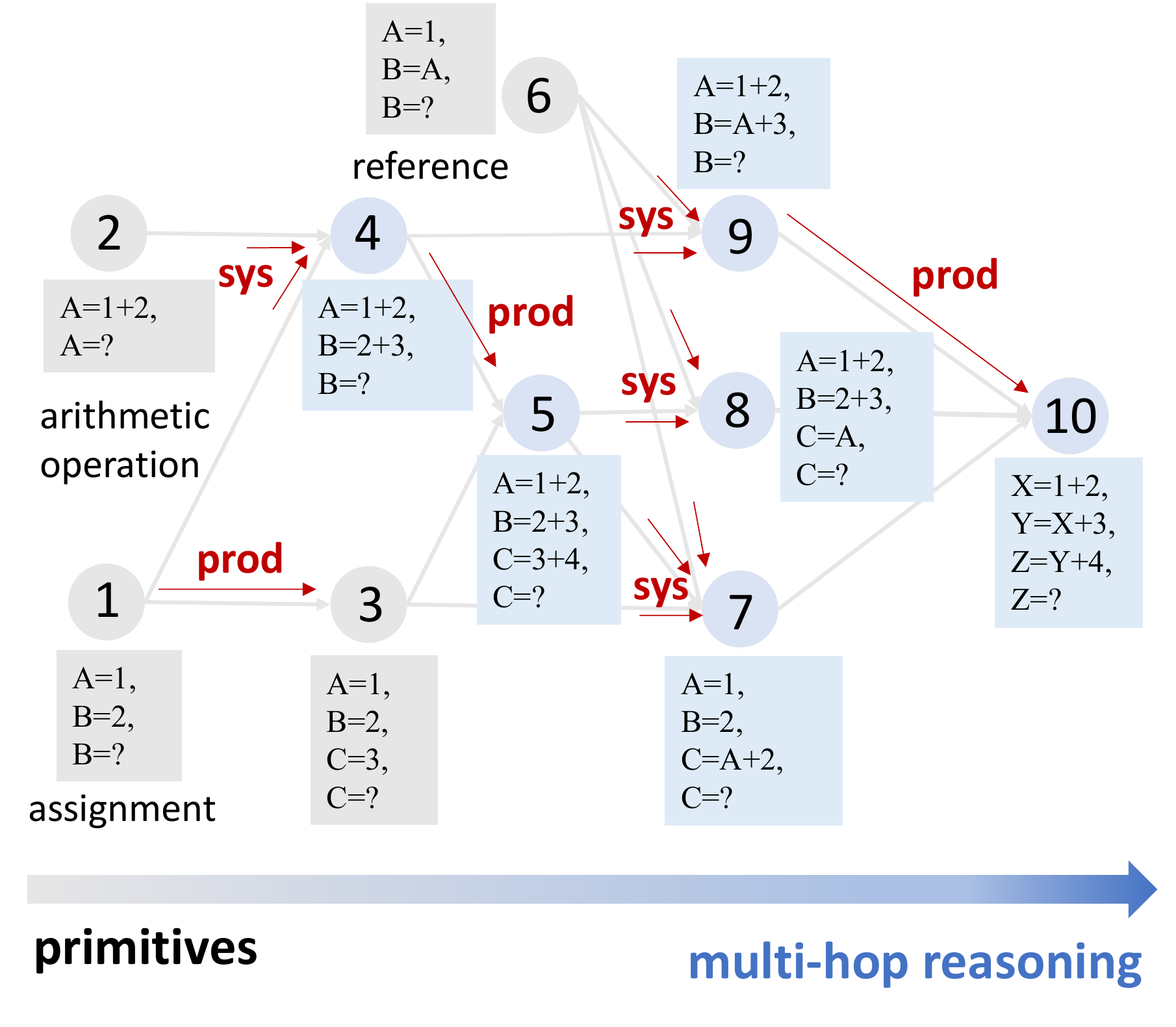}
\caption{Skill tree to evaluate compositional generalization. The data format of primitive operations is gray and others (complex formulas composed of combinations of primitive operations) are blue .}
\label{fig:skill-tree}
\end{figure}

Specifically, we designed ten versions of symbolic reasoning problems.
The hierarchical relationship of their task levels is illustrated in Figure~\ref{fig:skill-tree}; in this skill tree, each vertex, i.e., domain, corresponds to different task settings with different complexity, and edges represent the hierarchical complexity levels.

By adequately selecting a particular combination of training and test domains, we evaluated the compositional generalization ability of the models from various perspectives.
Here, the arithmetic expressions used in the test domain are a combination of those in training domains, creating a semi-order relationship in the skill tree.
For example, using the settings 1 (\texttt{A=1+2,A=?}) and 2 (\texttt{A=1,B=2,B=?}) as a training set, and 4 (\texttt{A=1+2,B=2+3,B=?}) as a test set, one can evaluate the model's systematicity generalization towards the arithmetic operations  (\texttt{a+b}) and assignments (\texttt{A=$i$,B=$j$,B=?}).

\begin{table}[!t]
\centering
\scriptsize

\begin{tabular}{llrrrrrr}
\toprule
& & \multicolumn{2}{c}{base} & \multicolumn{2}{c}{large} & \multicolumn{2}{c}{x-large} \\
Task & Type & ZA &     WA & ZA &     WA & ZA &     WA \\
\cmidrule(r){1-1} \cmidrule(lr){2-2} \cmidrule{3-8}
1,2         &     sys. &               \cellcolor[rgb]{0.8359073359073359, 0.95, 0.8359073359073359} $\!\!$42.0$\!\!$ &   \cellcolor[rgb]{0.8727445997458704, 0.8727445997458704, 0.95} $\!\!$82.1$\!\!$ &               \cellcolor[rgb]{0.8752895752895753, 0.95, 0.8752895752895753} $\!\!$35.2$\!\!$ &   \cellcolor[rgb]{0.8629606099110546, 0.8629606099110546, 0.95} $\!\!$89.8$\!\!$ &               \cellcolor[rgb]{0.7797297297297296, 0.95, 0.7797297297297296} $\!\!$51.7$\!\!$ &   \cellcolor[rgb]{0.8541931385006353, 0.8541931385006353, 0.95} $\!\!$96.7$\!\!$ \\
$\rightarrow$4         &  +subst. &               \cellcolor[rgb]{0.8527027027027027, 0.95, 0.8527027027027027} $\!\!$39.1$\!\!$ &   \cellcolor[rgb]{0.8749047013977128, 0.8749047013977128, 0.95} $\!\!$80.4$\!\!$ &               \cellcolor[rgb]{0.8868725868725869, 0.95, 0.8868725868725869} $\!\!$33.2$\!\!$ &   \cellcolor[rgb]{0.8634688691232528, 0.8634688691232528, 0.95} $\!\!$89.4$\!\!$ &               \cellcolor[rgb]{0.7855212355212355, 0.95, 0.7855212355212355} $\!\!$50.7$\!\!$ &   \cellcolor[rgb]{0.8541931385006353, 0.8541931385006353, 0.95} $\!\!$96.7$\!\!$ \\
\cmidrule(r){1-1} \cmidrule(lr){2-2} \cmidrule{3-8}
2,3         &     sys. &               \cellcolor[rgb]{0.8845559845559845, 0.95, 0.8845559845559845} $\!\!$33.6$\!\!$ &   \cellcolor[rgb]{0.8812579415501905, 0.8812579415501905, 0.95} $\!\!$75.4$\!\!$ &               \cellcolor[rgb]{0.8932432432432432, 0.95, 0.8932432432432432} $\!\!$32.1$\!\!$ &   \cellcolor[rgb]{0.8682973316391359, 0.8682973316391359, 0.95} $\!\!$85.6$\!\!$ &               \cellcolor[rgb]{0.8712355212355212, 0.95, 0.8712355212355212} $\!\!$35.9$\!\!$ &   \cellcolor[rgb]{0.8567344345616263, 0.8567344345616263, 0.95} $\!\!$94.7$\!\!$ \\
$\rightarrow$5         &  +subst. &               \cellcolor[rgb]{0.8845559845559845, 0.95, 0.8845559845559845} $\!\!$33.6$\!\!$ &   \cellcolor[rgb]{0.8792249047013977, 0.8792249047013977, 0.95} $\!\!$77.0$\!\!$ &               \cellcolor[rgb]{0.8967181467181466, 0.95, 0.8967181467181466} $\!\!$31.5$\!\!$ &   \cellcolor[rgb]{0.8662642947903431, 0.8662642947903431, 0.95} $\!\!$87.2$\!\!$ &               \cellcolor[rgb]{0.8677606177606177, 0.95, 0.8677606177606177} $\!\!$36.5$\!\!$ &   \cellcolor[rgb]{0.8564803049555273, 0.8564803049555273, 0.95} $\!\!$94.9$\!\!$ \\
\cmidrule(r){1-1} \cmidrule(lr){2-2} \cmidrule{3-8}
2,3,6         &     sys. &               \cellcolor[rgb]{0.8428571428571429, 0.95, 0.8428571428571429} $\!\!$40.8$\!\!$ &   \cellcolor[rgb]{0.8798602287166455, 0.8798602287166455, 0.95} $\!\!$76.5$\!\!$ &               \cellcolor[rgb]{0.8527027027027027, 0.95, 0.8527027027027027} $\!\!$39.1$\!\!$ &   \cellcolor[rgb]{0.8656289707750953, 0.8656289707750953, 0.95} $\!\!$87.7$\!\!$ &               \cellcolor[rgb]{0.8445945945945945, 0.95, 0.8445945945945945} $\!\!$40.5$\!\!$ &   \cellcolor[rgb]{0.8568614993646759, 0.8568614993646759, 0.95} $\!\!$94.6$\!\!$ \\
$\rightarrow$8         &  +subst. &               \cellcolor[rgb]{0.8515444015444016, 0.95, 0.8515444015444016} $\!\!$39.3$\!\!$ &   \cellcolor[rgb]{0.8821473951715374, 0.8821473951715374, 0.95} $\!\!$74.7$\!\!$ &               \cellcolor[rgb]{0.861969111969112, 0.95, 0.861969111969112} $\!\!$37.5$\!\!$ &   \cellcolor[rgb]{0.8672808132147395, 0.8672808132147395, 0.95} $\!\!$86.4$\!\!$ &               \cellcolor[rgb]{0.8521235521235521, 0.95, 0.8521235521235521} $\!\!$39.2$\!\!$ &   \cellcolor[rgb]{0.8564803049555273, 0.8564803049555273, 0.95} $\!\!$94.9$\!\!$ \\
\cmidrule(r){1-1} \cmidrule(lr){2-2} \cmidrule{3-8}
2,3,6         &     sys. &               \cellcolor[rgb]{0.7519305019305019, 0.95, 0.7519305019305019} $\!\!$56.5$\!\!$ &   \cellcolor[rgb]{0.8757941550190597, 0.8757941550190597, 0.95} $\!\!$79.7$\!\!$ &               \cellcolor[rgb]{0.7837837837837838, 0.95, 0.7837837837837838} $\!\!$51.0$\!\!$ &   \cellcolor[rgb]{0.8707115628970775, 0.8707115628970775, 0.95} $\!\!$83.7$\!\!$ &               \cellcolor[rgb]{0.7426640926640926, 0.95, 0.7426640926640926} $\!\!$58.1$\!\!$ &   \cellcolor[rgb]{0.8573697585768741, 0.8573697585768741, 0.95} $\!\!$94.2$\!\!$ \\
$\rightarrow$7         &  +subst. &               \cellcolor[rgb]{0.7455598455598456, 0.95, 0.7455598455598456} $\!\!$57.6$\!\!$ &   \cellcolor[rgb]{0.8751588310038119, 0.8751588310038119, 0.95} $\!\!$80.2$\!\!$ &               \cellcolor[rgb]{0.7832046332046332, 0.95, 0.7832046332046332} $\!\!$51.1$\!\!$ &   \cellcolor[rgb]{0.8680432020330369, 0.8680432020330369, 0.95} $\!\!$85.8$\!\!$ &               \cellcolor[rgb]{0.7507722007722007, 0.95, 0.7507722007722007} $\!\!$56.7$\!\!$ &   \cellcolor[rgb]{0.8562261753494282, 0.8562261753494282, 0.95} $\!\!$95.1$\!\!$ \\
\cmidrule(r){1-1} \cmidrule(lr){2-2} \cmidrule{3-8}
1,2,6         &     sys. &               \cellcolor[rgb]{0.9395752895752896, 0.95, 0.9395752895752896} $\!\!$24.1$\!\!$ &   \cellcolor[rgb]{0.9412325285895806, 0.9412325285895806, 0.95} $\!\!$28.2$\!\!$ &               \cellcolor[rgb]{0.9453667953667954, 0.95, 0.9453667953667954} $\!\!$23.1$\!\!$ &   \cellcolor[rgb]{0.9398348157560356, 0.9398348157560356, 0.95} $\!\!$29.3$\!\!$ &               \cellcolor[rgb]{0.9198841698841699, 0.95, 0.9198841698841699} $\!\!$27.5$\!\!$ &   \cellcolor[rgb]{0.9356416772554003, 0.9356416772554003, 0.95} $\!\!$32.6$\!\!$ \\
$\rightarrow$9         &  +subst. &               \cellcolor[rgb]{0.9297297297297297, 0.95, 0.9297297297297297} $\!\!$25.8$\!\!$ &   \cellcolor[rgb]{0.9414866581956798, 0.9414866581956798, 0.95} $\!\!$28.0$\!\!$ &               \cellcolor[rgb]{0.9395752895752896, 0.95, 0.9395752895752896} $\!\!$24.1$\!\!$ &   \cellcolor[rgb]{0.9367852604828462, 0.9367852604828462, 0.95} $\!\!$31.7$\!\!$ &               \cellcolor[rgb]{0.9140926640926641, 0.95, 0.9140926640926641} $\!\!$28.5$\!\!$ &   \cellcolor[rgb]{0.9329733163913595, 0.9329733163913595, 0.95} $\!\!$34.7$\!\!$ \\
\cmidrule(r){1-1} \cmidrule(lr){2-2} \cmidrule{3-8}
7,8 &     sys. &               \cellcolor[rgb]{0.9424710424710424, 0.95, 0.9424710424710424} $\!\!$23.6$\!\!$ &   \cellcolor[rgb]{0.95, 0.95, 0.95} $\!\!$21.3$\!\!$ &               \cellcolor[rgb]{0.9326254826254826, 0.95, 0.9326254826254826} $\!\!$25.3$\!\!$ &   \cellcolor[rgb]{0.9441550190597204, 0.9441550190597204, 0.95} $\!\!$25.9$\!\!$ &               \cellcolor[rgb]{0.95, 0.95, 0.95} $\!\!$22.3$\!\!$ &   \cellcolor[rgb]{0.9412325285895806, 0.9412325285895806, 0.95} $\!\!$28.2$\!\!$ \\
$\rightarrow$10 &  +subst. &               \cellcolor[rgb]{0.95, 0.95, 0.95} $\!\!$22.3$\!\!$ &   \cellcolor[rgb]{0.9496188055908513, 0.9496188055908513, 0.95} $\!\!$21.6$\!\!$ &               \cellcolor[rgb]{0.9378378378378378, 0.95, 0.9378378378378378} $\!\!$24.4$\!\!$ &   \cellcolor[rgb]{0.9435196950444726, 0.9435196950444726, 0.95} $\!\!$26.4$\!\!$ &               \cellcolor[rgb]{0.9459459459459459, 0.95, 0.9459459459459459} $\!\!$23.0$\!\!$ &   \cellcolor[rgb]{0.9386912325285895, 0.9386912325285895, 0.95} $\!\!$30.2$\!\!$ \\
\cmidrule(r){1-1} \cmidrule(lr){2-2} \cmidrule{3-8}
1         &    prod. &              \cellcolor[rgb]{0.5, 0.95, 0.5} $\!\!$100.0$\!\!$ &  \cellcolor[rgb]{0.85, 0.85, 0.95} $\!\!$100.0$\!\!$ &              \cellcolor[rgb]{0.5, 0.95, 0.5} $\!\!$100.0$\!\!$ &  \cellcolor[rgb]{0.85, 0.85, 0.95} $\!\!$100.0$\!\!$ &              \cellcolor[rgb]{0.5, 0.95, 0.5} $\!\!$100.0$\!\!$ &  \cellcolor[rgb]{0.85, 0.85, 0.95} $\!\!$100.0$\!\!$ \\
$\rightarrow$3         &  +subst. &              \cellcolor[rgb]{0.5, 0.95, 0.5} $\!\!$100.0$\!\!$ &  \cellcolor[rgb]{0.85, 0.85, 0.95} $\!\!$100.0$\!\!$ &              \cellcolor[rgb]{0.5, 0.95, 0.5} $\!\!$100.0$\!\!$ &  \cellcolor[rgb]{0.85, 0.85, 0.95} $\!\!$100.0$\!\!$ &              \cellcolor[rgb]{0.5, 0.95, 0.5} $\!\!$100.0$\!\!$ &  \cellcolor[rgb]{0.85, 0.85, 0.95} $\!\!$100.0$\!\!$ \\
\cmidrule(r){1-1} \cmidrule(lr){2-2} \cmidrule{3-8}
4         &    prod. &              \cellcolor[rgb]{0.5, 0.95, 0.5} $\!\!$100.0$\!\!$ &  \cellcolor[rgb]{0.85, 0.85, 0.95} $\!\!$100.0$\!\!$ &              \cellcolor[rgb]{0.5, 0.95, 0.5} $\!\!$100.0$\!\!$ &  \cellcolor[rgb]{0.85, 0.85, 0.95} $\!\!$100.0$\!\!$ &              \cellcolor[rgb]{0.5, 0.95, 0.5} $\!\!$100.0$\!\!$ &  \cellcolor[rgb]{0.85, 0.85, 0.95} $\!\!$100.0$\!\!$ \\
$\rightarrow$5         &  +subst. &               \cellcolor[rgb]{0.5005791505791506, 0.95, 0.5005791505791506} $\!\!$99.9$\!\!$ &   \cellcolor[rgb]{0.8501270648030496, 0.8501270648030496, 0.95} $\!\!$99.9$\!\!$ &              \cellcolor[rgb]{0.5, 0.95, 0.5} $\!\!$100.0$\!\!$ &  \cellcolor[rgb]{0.85, 0.85, 0.95} $\!\!$100.0$\!\!$ &              \cellcolor[rgb]{0.5, 0.95, 0.5} $\!\!$100.0$\!\!$ &   \cellcolor[rgb]{0.8501270648030496, 0.8501270648030496, 0.95} $\!\!$99.9$\!\!$ \\
\cmidrule(r){1-1} \cmidrule(lr){2-2} \cmidrule{3-8}
9         &    prod. &               \cellcolor[rgb]{0.749034749034749, 0.95, 0.749034749034749} $\!\!$57.0$\!\!$ &   \cellcolor[rgb]{0.9017153748411689, 0.9017153748411689, 0.95} $\!\!$59.3$\!\!$ &               \cellcolor[rgb]{0.7218146718146718, 0.95, 0.7218146718146718} $\!\!$61.7$\!\!$ &   \cellcolor[rgb]{0.895997458703939, 0.895997458703939, 0.95} $\!\!$63.8$\!\!$ &               \cellcolor[rgb]{0.7281853281853282, 0.95, 0.7281853281853282} $\!\!$60.6$\!\!$ &   \cellcolor[rgb]{0.897395171537484, 0.897395171537484, 0.95} $\!\!$62.7$\!\!$ \\
$\rightarrow$10         &  +subst. &               \cellcolor[rgb]{0.740926640926641, 0.95, 0.740926640926641} $\!\!$58.4$\!\!$ &   \cellcolor[rgb]{0.8996823379923761, 0.8996823379923761, 0.95} $\!\!$60.9$\!\!$ &               \cellcolor[rgb]{0.7189189189189189, 0.95, 0.7189189189189189} $\!\!$62.2$\!\!$ &   \cellcolor[rgb]{0.8956162642947904, 0.8956162642947904, 0.95} $\!\!$64.1$\!\!$ &               \cellcolor[rgb]{0.7345559845559845, 0.95, 0.7345559845559845} $\!\!$59.5$\!\!$ &   \cellcolor[rgb]{0.8951080050825921, 0.8951080050825921, 0.95} $\!\!$64.5$\!\!$ \\
\bottomrule
\end{tabular}

\caption{
Average accuracies in the experiment with 2 different seeds.
The ``Task'' column exhibits (train$\rightarrow$test) domains corresponding to the skill-tree (Figure~\ref{fig:skill-tree}).
The ``Type'' column shows the targeted compositionality type in each setting; here, ``sys.,'' ``prod,'' and ``subst.'' denote the systematicity, productivity, and substitutivity generalizations, respectively.  
}
\label{table:compositional_results}
\end{table}

\section{Experimental settings}
\label{sec:evaluation_method}

\subsection{Data}
\noindent
\textbf{Dataset:}
In each experimental setting, we refer to the training domains as $\mathcal D_\mathrm{train}=\{d_{\mathrm{train}1}, \cdots d_{\mathrm{train}k}\}$ and the test domain as $d_\mathrm{test}$.
Each domain has 100,000 training data and 3,200 test data; these are randomly generated, and there is no overlapped instance.
When the training domain consisted of multiple domains, we used the union of the training data in $\mathcal D_\mathrm{train}$.
In addition, when the training domain is not primitive operations (1, 2, 3, and 6 in Figure~\ref{fig:skill-tree}), we further added the primitive operation data related to the training domain (Appendix~\ref{sec:appendix:training_configurations}) into the training data.

\noindent
\textbf{Arithmetic expressions:}
As introduced in Section~\ref{sec:problem}, the input is a sequence of arithmetic expressions.
Formally, each expression is in the format of \texttt{a=$n$} or \texttt{a=$n$\{+,-,max,min\}$m$} except that the final expression asks the number assigned to a specified variable (\texttt{b=?}).
Here, \texttt{a} and \texttt{b} are a member of a variable name set $\Sigma$; $n$ and $m$ are a member of the variable name set or number set $\Sigma \cup \mathcal N$.
Specifically, $\Sigma$ consists of 21 alphabets, and $N$ consists of the integer from 0 to 99.
The symbol \texttt{=} indicates that the result of the left-hand side is substituted into the right-hand side.
The operations (\texttt{+}, \texttt{-}, \texttt{max}, and \texttt{min}) correspond to arithmetic addition, subtraction, max (returning the larger of its left and right numbers), and min (returning the smaller of its left and right numbers).

The questions are designed so that the answer is unique, and depending on the problem set-up, may include mathematical expressions that are not directly related to the final answer, i.e., distractors.
The order of the equations is arbitrary; the first equation should not necessarily be calculated first.

\noindent
\textbf{Substitivity test:}
In each experimental setting, we evaluate the substitutivity generalization performance of the model under the situation where the variable names are replaced with unseen ones, e.g., training with \texttt{a=1+2,a=?}; then evaluating with \texttt{α=2+4,α=?}.
In this setting, we replaced each variable name in the test set with one of five alphabets that do not overlap with the training ones.

\subsection{Trainig and test}

\noindent
\textbf{Training:}
The training stops when the accuracy on the validation dataset does not increase in successive five epochs or until the validation accuracy reaches 100\%. 
Checkpoints with the highest accuracy in the validation dataset are used for evaluation. 
Note that among the experiments, the accuracy in the training domain reached at least 99.5\%; this indicates that the primitive operations were learnable for the models.
Detailed settings for training are described in Appendix~\ref{sec:appendix:training_configurations}.

\noindent
\textbf{Evaluation metrics:}
The accuracy is calculated by the test data in the test domain $d_\mathrm{test}$.
Here, we used two metrics: (i) zero-shot accuracy (ZA) and (ii) weighted average of accuracies (WA) to measure the efficiency of learning~\cite{talmor-etal-2020-olmpics}.
In measuring WA, a model was further trained using the training set in the test domain $d_\mathrm{test}$; then, the weighted average of accuracies at every update was calculated (details are in Appendix~\ref{sec:appendix:how_to_calculate_ws}).

\subsection{Models:}
We used three different sizes (base, large, and xl) of T5~\cite{JMLR:v21:20-074}, which is a widely used pre-trained seq2seq model in numerical reasoning tasks~\cite{pal-baral-2021-investigating-numeracy, DBLP:journals/corr/abs-2210-11416, DBLP:journals/corr/abs-2104-07307}.
Note that we began our training using the models with learned parameters.
We also evaluated BART variants and randomly initialized models in Appendix~\ref{sec:appendix:models}.

\section{Experiments and results}
\label{section:results}
We adopted nine combinations of training and test domains as shown in the first column of Table~\ref{table:compositional_results} (training domains$\rightarrow$test domain).
Six of them test the systematicity generalization and the other three test productivity generalization.
In each setting, we further tested substitutivity generalization ability using the test domain data with a different variable name set (e.g., \texttt{α} instead of \texttt{A}) to that used in the training domain.
Table~\ref{table:compositional_results} shows the overall results.
We observed the following four trends:
\begin{itemize}
  \setlength{\parskip}{0cm} %
  \setlength{\itemsep}{0cm} %
    \item Systematicity generalization was more difficult than productivity generalization.
    \item Even in the simple composition (the setting 1,2$\rightarrow$4), the models struggle with generalization from zero or few examples.
    \item Models achieved substitutivity generalization.
    \item Model size did not incur substantial performance difference.
\end{itemize}

We identified that the systematicity generalization of reference and arithmetic operations (setting $2,3\rightarrow5$; from \texttt{A=1,B=2,C=3,C=?} and \texttt{A=1+2,A=?} to \texttt{A=1+2,B=2+3,C=4+5,C=?}) was a simple setting, yet difficult to solve (refer to Appendix~\ref{appendix:sec:detailed_analysis_of_t5_based_models} for results on other tasks.).
To better understand why neural models struggle with this setting, we decomposed the complexity of this setting and analyzed the model performance.
Note that \citet{kim-linzen-2020-cogs} also suggested that neural models lack systematicity generalization ability in the context of semantic parsing; our results corroborate their findings from the context of arithmetic multi-hop reasoning.

\begin{table}[!t]
\centering
\small

\begin{tabular}{lrrrrrrr}
\toprule
& \multicolumn{2}{c}{base} & \multicolumn{2}{c}{large} & \multicolumn{2}{c}{x-large} \\
Setting & ZA &     WA & ZA &     WA & ZA &     WA \\
\cmidrule(r){1-1} \cmidrule{2-7}
2,3$\rightarrow$5     &                 \cellcolor[rgb]{0.8951400329489291, 0.95, 0.8951400329489291} $\!\!$33.6$\!\!$ &  \cellcolor[rgb]{0.95, 0.95, 0.95} $\!\!$75.4$\!\!$ &               \cellcolor[rgb]{0.9062602965403623, 0.95, 0.9062602965403623} $\!\!$32.1$\!\!$ &  \cellcolor[rgb]{0.9073221757322176, 0.9073221757322176, 0.95} $\!\!$85.6$\!\!$ &               \cellcolor[rgb]{0.8780889621087314, 0.95, 0.8780889621087314} $\!\!$35.9$\!\!$ &  \cellcolor[rgb]{0.8692468619246861, 0.8692468619246861, 0.95} $\!\!$94.7$\!\!$ \\
\cmidrule(r){1-1} \cmidrule{2-7}
String     &                 \cellcolor[rgb]{0.8677100494233937, 0.95, 0.8677100494233937} $\!\!$37.3$\!\!$ &  \cellcolor[rgb]{0.8717573221757322, 0.8717573221757322, 0.95} $\!\!$94.1$\!\!$ &               \cellcolor[rgb]{0.6542009884678748, 0.95, 0.6542009884678748} $\!\!$66.1$\!\!$ &  \cellcolor[rgb]{0.853765690376569, 0.853765690376569, 0.95} $\!\!$98.4$\!\!$ &               \cellcolor[rgb]{0.5, 0.95, 0.5} $\!\!$86.9$\!\!$ &  \cellcolor[rgb]{0.85, 0.85, 0.95} $\!\!$99.3$\!\!$ \\
Steps &                 \cellcolor[rgb]{0.95, 0.95, 0.95} $\!\!$26.2$\!\!$ &  \cellcolor[rgb]{0.9219665271966527, 0.9219665271966527, 0.95} $\!\!$82.1$\!\!$ &               \cellcolor[rgb]{0.8766062602965403, 0.95, 0.8766062602965403} $\!\!$36.1$\!\!$ &  \cellcolor[rgb]{0.8914225941422593, 0.8914225941422593, 0.95} $\!\!$89.4$\!\!$ &               \cellcolor[rgb]{0.8943986820428336, 0.95, 0.8943986820428336} $\!\!$33.7$\!\!$ &  \cellcolor[rgb]{0.8621338912133891, 0.8621338912133891, 0.95} $\!\!$96.4$\!\!$ \\
\bottomrule
\end{tabular}

\caption{
Ablation study with the 2,3$\rightarrow$5 (vanilla) setting.
``String'' refers to the setting where string operations are used instead of arithmetic operations.
``Step'' denotes the setting generating intermediate steps.
}
\label{table:ablation_study_results}
\end{table}

\paragraph{Is this difficulty specific to \textit{arithmetic} symbolic reasoning?}
We experimented with the same setting except that the four arithmetic operations are replaced with string operations (\texttt{join}, \texttt{reserveJoin}, \texttt{strSub}, and \texttt{stackJoin}; details are in Appendix~\ref{appendix:subsec:straing_operations}).
The notable difference between arithmetic and string operations is that the string operation could be achieved by only copying selective elements in the input (e.g., \texttt{12+34=1234}), while arithmetic operation requires the models to access the arithmetic knowledge stored in their internals (e.g., \texttt{1+2=3}) and generate new information not stated in the input context (e.g., 3).

Larger models tended to overcome the weakness in composition with string operations (e.g., the accuracy of 86.9 in zero-shot evaluation with the x-large model), while they struggled with arithmetic operations.
This suggests that the major difficulty in systematicity was in \textbf{the access to the arithmetic knowledge (e.g., 1+1=2)}.

\paragraph{Does scratchpad training alleviate the difficulty?}
Existing studies suggested that showing the intermediate step (scratchpad-style training/inference) improves the multi-hop reasoning ability of neural models~\cite{wei2022chain}.
We tested whether such an explicit generation of intermediate information alleviates the difficulty faced in the previous analysis.
Specifically, we trained models with intermediate steps (e,g. \texttt{A=1+2,B=2+3,B=?; B=2+3,B=5}. Details are in Appendix~\ref{appendix:subsec:scratch_pad_formulation}) during training.

The accuracy was calculated by the exact match of the answer and intermediate steps (the steps are designed to be uniquely determined).
The performance gain due to explicating the intermediate steps was limited (Table~\ref{table:ablation_study_results}), at least with our T5-based models.
This shows that, in our carefully controlled setting, merely employing the scratchpad-style generation is not substantially effective.

\begin{table*}[!t]
\centering

\begin{tabular}{lrrrrrrr}
\toprule
Complexity & \multicolumn{2}{c}{base} & \multicolumn{2}{c}{large} & \multicolumn{2}{c}{x-large} & \multirow{2}{*}{Average} \\
 dimensions & ZA &  WA & ZA & WA & ZA & WA &  \\
\midrule
 $\Delta$\#variables            &    0.098 &   -0.098 &     0.488 &    -0.293 &      -0.098 &      -0.488 &   -0.065 \\
 $\Delta$\#numbers                 &    0.059 &    0.265 &    -0.088 &     0.647 &       0.206 &       0.677 &    0.294 \\
 $\Delta$\#operations           &   -0.507 &   -0.338 &    -0.338 &     0.169 &      -0.338 &       0.169 &   -0.197 \\
 $\Delta$\#reference &   -0.655 &   -0.655 &    -0.393 &    -0.655 &      -0.655 &      -0.655 &   \textbf{-0.611} \\
\bottomrule
\end{tabular}

\caption{
Spearman's rank correlation coefficient between the increase of training--test arithmetic complexity and the compositional generalization performance (accuracy) across the nine settings listed in Table~\ref{table:compositional_results}.
A negative score indicates that the greater the training--test discrepancy in its dimension, the more difficult compositional generalization is.
In the case that there are multiple training domains, the maximum value among them is used.
}
\label{table:correlation_analysis}
\end{table*}

\section{Analysis}
We conduct a more in-depth analysis of compositional generalization difficulties from another perspective---complexity of arithmetic expressions.
Specifically, for each pair of training and test domains listed in Table~\ref{table:compositional_results} (e.g., 1,2$\rightarrow$4), we quantified the increase of the complexity of arithmetic formulas from several aspects, e.g., how much the formula's number of variables increased in the test domain (setting of 4) compared to the training domain (setting of 1 and 2).
Specifically, we focused on the increase of the number of variables ($\Delta$\#variables), numbers ($\Delta$\#numbers), operations ($\Delta$\#operations), and references ($\Delta$\#references) from the training to test domains.
Here, ``\#reference'' denotes the number of access to a particular variable on the right-hand of equations.
For example, the $\Delta$\#references is 1 if the training data format is \texttt{A=$n$,B=\textbf{A}+$m$} and the test format is \texttt{A=$n$,B=\textbf{A}+$m$,C=\textbf{B}+$l$}.
Then, we identified which dimension strongly relates to the compositional generalization difficulty.

We analyzed the macro trends between formula's complexity increase and the difficulty of generalization across the experimental settings.
Table~\ref{table:correlation_analysis} shows Spearman's rank correlation coefficient between each complexity and the test-domain accuracy.
We found a notable negative correlation in the $\Delta$\#reference; that is, the more references in the test domain compared to the training domain, the more difficult the compositional generalization becomes (the cases of 1,2,6$\rightarrow$9 and 9$\rightarrow$10 settings).
Simply put, this reveals the difficulty of compositional generalization with \textit{multi-hop} reasoning---retaining the results of a calculation and accessing them again for another calculation.

\section{Related work}
The analysis of the compositional generalization ability of neural models and arithmetic multi-hop reasoning problems have typically been studied separately; this study has merged these two directions.
As for composition generalization analysis, several studies analyzed neural models using datasets such as SCAN~\cite{Lake2018GeneralizationWS}, COGS~\cite{kim-linzen-2020-cogs}, and CFQ~\cite{keysers2020measuring}.
These mainly focused on compositionality in the context of semantic parsing; the composition ability toward symbol manipulations (e.g., multi-hop arithmetic reasoning) is typically out of focus.
As for arithmetic reasoning, neural models' abilities have been analyzed typically using benchmarks such as DROP~\cite{dua-etal-2019-drop}.
It has recently been reported that such dataset has superficial cues~\cite{al-negheimish-etal-2021-numerical}, which made it unclear how much arithmetic reasoning neural model achieves; our study using a carefully controlled dataset contributed to the exact weakness of neural models in this context.

\section{Conclusion}
\label{sec:conclusion}
In this study, we have empirically investigated the arithmetic multi-hop reasoning ability of modern neural models through the lens of compositional generalization ability.
To systematically analyze neural models' ability, we have defined a skill tree that organizes the (hierarchical) complexity levels of the multi-hop symbolic reasoning dataset.

Our experiments have revealed that the major weakness lies in systematicity, even with a relatively simple composition.
Through the ablation studies, we also have found that difficulty in systematicity is pronounced in accessing knowledge that is not written in input but stored in models.
Furthermore, even in training models with intermediate steps that explicate the composition, they struggle to capture systematicity.
We also found the difficulty of multi-hop reasoning in compositional generalization.
These highlight the exact weakness of neural models and encourage studies to overcome such limitations.

\section*{Limitations}
In this work, we explored neural networks' ability to capture compositionality in symbolic arithmetic reasoning in hopes that it may lead to future improvements in more general reasoning. 
However, arithmetic reasoning may not necessarily generalize to natural language tasks.
Furthermore, we explored several aspects of multi-hop arithmetic reasoning, but these were chosen from a relatively human-centric perspective, and models may suffer from unforeseen other difficulties. 
Finally, while we found several patterns in how model performance degrades, it is difficult to aggregate this into a full picture of what a model can and cannot do. 
Further experiments are needed to gain a more complete understanding of model performance

\section*{Acknowledgements}
We thank four anonymous reviewers who provided valuable feedback.
We would like to also appreciate the member of Tohoku NLP Group for their cooperation in conducting this research.
This work was supported by JST CREST JPMJCR20D2 and JSPS KAKENHI Grant Number JP22H00524，21K21343.

\bibliography{anthology,custom}

\begin{thebibliography}{21}
\expandafter\ifx\csname natexlab\endcsname\relax\def\natexlab#1{#1}\fi

\bibitem[{Al-Negheimish et~al.(2021)Al-Negheimish, Madhyastha, and
  Russo}]{al-negheimish-etal-2021-numerical}
Hadeel Al-Negheimish, Pranava Madhyastha, and Alessandra Russo. 2021.
\newblock \href {https://doi.org/10.18653/v1/2021.emnlp-main.759} {Numerical
  reasoning in machine reading comprehension tasks: are we there yet?}
\newblock In \emph{Proceedings of the 2021 Conference on Empirical Methods in
  Natural Language Processing}, pages 9643--9649, Online and Punta Cana,
  Dominican Republic. Association for Computational Linguistics.

\bibitem[{Chung et~al.(2022)Chung, Hou, Longpre, Zoph, Tay, Fedus, Li, Wang,
  Dehghani, Brahma, Webson, Gu, Dai, Suzgun, Chen, Chowdhery, Narang, Mishra,
  Yu, Zhao, Huang, Dai, Yu, Petrov, Chi, Dean, Devlin, Roberts, Zhou, Le, and
  Wei}]{DBLP:journals/corr/abs-2210-11416}
Hyung~Won Chung, Le~Hou, Shayne Longpre, Barret Zoph, Yi~Tay, William Fedus,
  Eric Li, Xuezhi Wang, Mostafa Dehghani, Siddhartha Brahma, Albert Webson,
  Shixiang~Shane Gu, Zhuyun Dai, Mirac Suzgun, Xinyun Chen, Aakanksha
  Chowdhery, Sharan Narang, Gaurav Mishra, Adams Yu, Vincent~Y. Zhao, Yanping
  Huang, Andrew~M. Dai, Hongkun Yu, Slav Petrov, Ed~H. Chi, Jeff Dean, Jacob
  Devlin, Adam Roberts, Denny Zhou, Quoc~V. Le, and Jason Wei. 2022.
\newblock \href {https://doi.org/10.48550/arXiv.2210.11416} {Scaling
  instruction-finetuned language models}.
\newblock \emph{CoRR}, abs/2210.11416.

\bibitem[{Clark et~al.(2020)Clark, Tafjord, and Richardson}]{ijcai2020p537}
Peter Clark, Oyvind Tafjord, and Kyle Richardson. 2020.
\newblock \href {https://doi.org/10.24963/ijcai.2020/537} {Transformers as soft
  reasoners over language}.
\newblock In \emph{Proceedings of the Twenty-Ninth International Joint
  Conference on Artificial Intelligence, {IJCAI-20}}, pages 3882--3890.
  International Joint Conferences on Artificial Intelligence Organization.
\newblock Main track.

\bibitem[{d'Avila Garcez and Lamb(2020)}]{DBLP:journals/corr/abs-2012-05876}
Artur~S. d'Avila Garcez and Lu{\'{\i}}s~C. Lamb. 2020.
\newblock \href {http://arxiv.org/abs/2012.05876} {Neurosymbolic {AI:} the 3rd
  wave}.
\newblock \emph{CoRR}, abs/2012.05876.

\bibitem[{Dua et~al.(2019)Dua, Wang, Dasigi, Stanovsky, Singh, and
  Gardner}]{dua-etal-2019-drop}
Dheeru Dua, Yizhong Wang, Pradeep Dasigi, Gabriel Stanovsky, Sameer Singh, and
  Matt Gardner. 2019.
\newblock \href {https://doi.org/10.18653/v1/N19-1246} {{DROP}: A reading
  comprehension benchmark requiring discrete reasoning over paragraphs}.
\newblock In \emph{Proceedings of the 2019 Conference of the North {A}merican
  Chapter of the Association for Computational Linguistics: Human Language
  Technologies, Volume 1 (Long and Short Papers)}, pages 2368--2378,
  Minneapolis, Minnesota. Association for Computational Linguistics.

\bibitem[{Geva et~al.(2020)Geva, Gupta, and Berant}]{geva-etal-2020-injecting}
Mor Geva, Ankit Gupta, and Jonathan Berant. 2020.
\newblock \href {https://doi.org/10.18653/v1/2020.acl-main.89} {Injecting
  numerical reasoning skills into language models}.
\newblock In \emph{Proceedings of the 58th Annual Meeting of the Association
  for Computational Linguistics}, pages 946--958, Online. Association for
  Computational Linguistics.

\bibitem[{Hupkes et~al.(2020)Hupkes, Dankers, Mul, and Bruni}]{ijcai2020p708}
Dieuwke Hupkes, Verna Dankers, Mathijs Mul, and Elia Bruni. 2020.
\newblock \href {https://doi.org/10.24963/ijcai.2020/708} {Compositionality
  decomposed: How do neural networks generalise? (extended abstract)}.
\newblock In \emph{Proceedings of the Twenty-Ninth International Joint
  Conference on Artificial Intelligence, {IJCAI-20}}, pages 5065--5069.
  International Joint Conferences on Artificial Intelligence Organization.
\newblock Journal track.

\bibitem[{Keysers et~al.(2020)Keysers, Sch{\"a}rli, Scales, Buisman, Furrer,
  Kashubin, Momchev, Sinopalnikov, Stafiniak, Tihon, Tsarkov, Wang, van Zee,
  and Bousquet}]{keysers2020measuring}
Daniel Keysers, Nathanael Sch{\"a}rli, Nathan Scales, Hylke Buisman, Daniel
  Furrer, Sergii Kashubin, Nikola Momchev, Danila Sinopalnikov, Lukasz
  Stafiniak, Tibor Tihon, Dmitry Tsarkov, Xiao Wang, Marc van Zee, and Olivier
  Bousquet. 2020.
\newblock \href {https://openreview.net/forum?id=SygcCnNKwr} {Measuring
  compositional generalization: A comprehensive method on realistic data}.
\newblock In \emph{International Conference on Learning Representations}.

\bibitem[{Kim and Linzen(2020)}]{kim-linzen-2020-cogs}
Najoung Kim and Tal Linzen. 2020.
\newblock \href {https://doi.org/10.18653/v1/2020.emnlp-main.731} {{COGS}: A
  compositional generalization challenge based on semantic interpretation}.
\newblock In \emph{Proceedings of the 2020 Conference on Empirical Methods in
  Natural Language Processing (EMNLP)}, pages 9087--9105, Online. Association
  for Computational Linguistics.

\bibitem[{Lake and Baroni(2018)}]{Lake2018GeneralizationWS}
Brenden~M. Lake and Marco Baroni. 2018.
\newblock Generalization without systematicity: On the compositional skills of
  sequence-to-sequence recurrent networks.
\newblock In \emph{ICML}.

\bibitem[{Lewis et~al.(2020)Lewis, Liu, Goyal, Ghazvininejad, Mohamed, Levy,
  Stoyanov, and Zettlemoyer}]{lewis-etal-2020-bart}
Mike Lewis, Yinhan Liu, Naman Goyal, Marjan Ghazvininejad, Abdelrahman Mohamed,
  Omer Levy, Veselin Stoyanov, and Luke Zettlemoyer. 2020.
\newblock \href {https://doi.org/10.18653/v1/2020.acl-main.703} {{BART}:
  Denoising sequence-to-sequence pre-training for natural language generation,
  translation, and comprehension}.
\newblock In \emph{Proceedings of the 58th Annual Meeting of the Association
  for Computational Linguistics}, pages 7871--7880, Online. Association for
  Computational Linguistics.

\bibitem[{Marcus(2003)}]{marcus2003algebraic}
Gary~F Marcus. 2003.
\newblock \emph{The algebraic mind: Integrating connectionism and cognitive
  science}.
\newblock MIT press.

\bibitem[{Pal and Baral(2021)}]{pal-baral-2021-investigating-numeracy}
Kuntal~Kumar Pal and Chitta Baral. 2021.
\newblock \href {https://doi.org/10.18653/v1/2021.findings-emnlp.265}
  {Investigating numeracy learning ability of a text-to-text transfer model}.
\newblock In \emph{Findings of the Association for Computational Linguistics:
  EMNLP 2021}, pages 3095--3101, Punta Cana, Dominican Republic. Association
  for Computational Linguistics.

\bibitem[{Qian et~al.(2022)Qian, Wang, Li, Li, and
  Yan}]{DBLP:journals/corr/abs-2208-05051}
Jing Qian, Hong Wang, Zekun Li, Shiyang Li, and Xifeng Yan. 2022.
\newblock \href {https://doi.org/10.48550/arXiv.2208.05051} {Limitations of
  language models in arithmetic and symbolic induction}.
\newblock \emph{CoRR}, abs/2208.05051.

\bibitem[{Rae et~al.(2021)Rae, Borgeaud, Cai, Millican, Hoffmann, Song,
  Aslanides, Henderson, Ring, Young, Rutherford, Hennigan, Menick, Cassirer,
  Powell, van~den Driessche, Hendricks, Rauh, Huang, Glaese, Welbl, Dathathri,
  Huang, Uesato, Mellor, Higgins, Creswell, McAleese, Wu, Elsen, Jayakumar,
  Buchatskaya, Budden, Sutherland, Simonyan, Paganini, Sifre, Martens, Li,
  Kuncoro, Nematzadeh, Gribovskaya, Donato, Lazaridou, Mensch, Lespiau,
  Tsimpoukelli, Grigorev, Fritz, Sottiaux, Pajarskas, Pohlen, Gong, Toyama,
  de~Masson~d'Autume, Li, Terzi, Mikulik, Babuschkin, Clark, de~Las~Casas, Guy,
  Jones, Bradbury, Johnson, Hechtman, Weidinger, Gabriel, Isaac, Lockhart,
  Osindero, Rimell, Dyer, Vinyals, Ayoub, Stanway, Bennett, Hassabis,
  Kavukcuoglu, and Irving}]{DBLP:journals/corr/abs-2112-11446}
Jack~W. Rae, Sebastian Borgeaud, Trevor Cai, Katie Millican, Jordan Hoffmann,
  H.~Francis Song, John Aslanides, Sarah Henderson, Roman Ring, Susannah Young,
  Eliza Rutherford, Tom Hennigan, Jacob Menick, Albin Cassirer, Richard Powell,
  George van~den Driessche, Lisa~Anne Hendricks, Maribeth Rauh, Po{-}Sen Huang,
  Amelia Glaese, Johannes Welbl, Sumanth Dathathri, Saffron Huang, Jonathan
  Uesato, John Mellor, Irina Higgins, Antonia Creswell, Nat McAleese, Amy Wu,
  Erich Elsen, Siddhant~M. Jayakumar, Elena Buchatskaya, David Budden, Esme
  Sutherland, Karen Simonyan, Michela Paganini, Laurent Sifre, Lena Martens,
  Xiang~Lorraine Li, Adhiguna Kuncoro, Aida Nematzadeh, Elena Gribovskaya,
  Domenic Donato, Angeliki Lazaridou, Arthur Mensch, Jean{-}Baptiste Lespiau,
  Maria Tsimpoukelli, Nikolai Grigorev, Doug Fritz, Thibault Sottiaux, Mantas
  Pajarskas, Toby Pohlen, Zhitao Gong, Daniel Toyama, Cyprien
  de~Masson~d'Autume, Yujia Li, Tayfun Terzi, Vladimir Mikulik, Igor
  Babuschkin, Aidan Clark, Diego de~Las~Casas, Aurelia Guy, Chris Jones, James
  Bradbury, Matthew~J. Johnson, Blake~A. Hechtman, Laura Weidinger, Iason
  Gabriel, William~S. Isaac, Edward Lockhart, Simon Osindero, Laura Rimell,
  Chris Dyer, Oriol Vinyals, Kareem Ayoub, Jeff Stanway, Lorrayne Bennett,
  Demis Hassabis, Koray Kavukcuoglu, and Geoffrey Irving. 2021.
\newblock \href {http://arxiv.org/abs/2112.11446} {Scaling language models:
  Methods, analysis {\&} insights from training gopher}.
\newblock \emph{CoRR}, abs/2112.11446.

\bibitem[{Raffel et~al.(2020)Raffel, Shazeer, Roberts, Lee, Narang, Matena,
  Zhou, Li, and Liu}]{JMLR:v21:20-074}
Colin Raffel, Noam Shazeer, Adam Roberts, Katherine Lee, Sharan Narang, Michael
  Matena, Yanqi Zhou, Wei Li, and Peter~J. Liu. 2020.
\newblock \href {http://jmlr.org/papers/v21/20-074.html} {Exploring the limits
  of transfer learning with a unified text-to-text transformer}.
\newblock \emph{Journal of Machine Learning Research}, 21(140):1--67.

\bibitem[{Shazeer and Stern(2018)}]{DBLP:conf/icml/ShazeerS18}
Noam Shazeer and Mitchell Stern. 2018.
\newblock \href {http://proceedings.mlr.press/v80/shazeer18a.html} {Adafactor:
  Adaptive learning rates with sublinear memory cost}.
\newblock In \emph{Proceedings of the 35th International Conference on Machine
  Learning, {ICML} 2018, Stockholmsm{\"{a}}ssan, Stockholm, Sweden, July 10-15,
  2018}, volume~80 of \emph{Proceedings of Machine Learning Research}, pages
  4603--4611. {PMLR}.

\bibitem[{Talmor et~al.(2020)Talmor, Elazar, Goldberg, and
  Berant}]{talmor-etal-2020-olmpics}
Alon Talmor, Yanai Elazar, Yoav Goldberg, and Jonathan Berant. 2020.
\newblock \href {https://doi.org/10.1162/tacl_a_00342} {o{LM}pics-on what
  language model pre-training captures}.
\newblock \emph{Transactions of the Association for Computational Linguistics},
  8:743--758.

\bibitem[{Tondello and Nacke(2019)}]{22136}
Gustavo~F. Tondello and Lennart~E. Nacke. 2019.
\newblock \href {http://ceur-ws.org/Vol-2497/paper15.pdf} {A pilot study of a
  digital skill tree in gameful education}.
\newblock In \emph{Proceedings of the 3rd International Symposium on
  Gamification and Games for Learning - GamiLearn {\textquoteright}19}.
  CEUR-WS.org, CEUR-WS.org.

\bibitem[{Wei et~al.(2022)Wei, Wang, Schuurmans, Bosma, brian ichter, Xia, Chi,
  Le, and Zhou}]{wei2022chain}
Jason Wei, Xuezhi Wang, Dale Schuurmans, Maarten Bosma, brian ichter, Fei Xia,
  Ed~H. Chi, Quoc~V Le, and Denny Zhou. 2022.
\newblock \href {https://openreview.net/forum?id=_VjQlMeSB_J} {Chain of thought
  prompting elicits reasoning in large language models}.
\newblock In \emph{Advances in Neural Information Processing Systems}.

\bibitem[{Yang et~al.(2021)Yang, Chen, Chen, and
  Cer}]{DBLP:journals/corr/abs-2104-07307}
Peng{-}Jian Yang, Ying{-}Ting Chen, Yuechan Chen, and Daniel Cer. 2021.
\newblock \href {http://arxiv.org/abs/2104.07307} {Nt5?! training {T5} to
  perform numerical reasoning}.
\newblock \emph{CoRR}, abs/2104.07307.

\end{thebibliography}
\bibliographystyle{acl_natbib}

\clearpage
\appendix

\section{Training configurations}
\label{sec:appendix:training_configurations}

\paragraph{Training data:}
Data of primitive domains (grey domains in Figure~\ref{fig:skill-tree}) are also added to the training data.
Specifically, data in the primitive domains that are reached by traversing the graph (Figure~\ref{fig:skill-tree}) from the training domain to the left is added. 
For example, when the training domain has the domain of 7, the data of primitive domains of 2, 3, and 6 are added\footnote{Task 7,8$\rightarrow$10 do not incorporate data from the primitive domains into the training data to evaluate if compositional generalization can be achieved solely from complex components.}.
Note that when the domain of 3 is included, the domain of 1 is not added.
Additionally, as the objective of this study does not emphasize the generalization performance in arithmetic ability, the scope of numbers utilized in the test domain is adjusted to ensure that the upper limit of the answers in the test domain does not surpass that of the answers in the train domain.

\paragraph{Hyperparameter:}
We used the T5 models (v1.1) as pre-trained model\footnote{\url{https://huggingface.co/docs/transformers/model_doc/t5v1.1}}.
We used three model sizes: base (250 million parameters), large (800 million parameters), and xl (3 billion parameters).
Following T5~\cite{JMLR:v21:20-074} fine-tuning configurations, we use Adafactor~\cite{DBLP:conf/icml/ShazeerS18} as the optimizer with a constant learning rate.
Also, we specify the learning rate  $1.0\times10^{-5}$ for training and $5.0\times10^{-5}$ in measuring the WA.
The batch size is 32 for all experimental settings. 
We trained each model on NVIDIA A6000 (48GB memory), A100 (80GB memory).
In addition, following previous research about numerical reasoning~\cite{geva-etal-2020-injecting}, we tokenize numbers in a digit-by-digit manner.

\begin{table}[t]
\centering
\begin{tabular}{llrrrr}
\toprule
& & \multicolumn{2}{c}{base} & \multicolumn{2}{c}{large} \\
Task & Type & ZA &     WA & ZA &     WA \\
\cmidrule(r){1-1} \cmidrule(lr){2-2} \cmidrule{3-6}
1,2         &     sys. &               \cellcolor[rgb]{0.9249049429657794, 0.95, 0.9249049429657794} $\!\!$25.5$\!\!$ &  \cellcolor[rgb]{0.8841696535244922, 0.8841696535244922, 0.95} $\!\!$71.2$\!\!$ &               \cellcolor[rgb]{0.8775665399239544, 0.95, 0.8775665399239544} $\!\!$33.8$\!\!$ &  \cellcolor[rgb]{0.8963560334528076, 0.8963560334528076, 0.95} $\!\!$61.0$\!\!$ \\
$\rightarrow$4         &  +subst. &               \cellcolor[rgb]{0.9266159695817491, 0.95, 0.9266159695817491} $\!\!$25.2$\!\!$ &  
\cellcolor[rgb]{0.8827359617682198, 0.8827359617682198, 0.95} $\!\!$72.4$\!\!$ &               \cellcolor[rgb]{0.8941064638783269, 0.95, 0.8941064638783269} $\!\!$30.9$\!\!$ &  \cellcolor[rgb]{0.8951612903225806, 0.8951612903225806, 0.95} $\!\!$62.0$\!\!$ \\
\cmidrule(r){1-1} \cmidrule(lr){2-2}  \cmidrule{3-6}
2,3         &     sys. &               \cellcolor[rgb]{0.9391634980988592, 0.95, 0.9391634980988592} $\!\!$23.0$\!\!$ &  \cellcolor[rgb]{0.9110513739545997, 0.9110513739545997, 0.95} $\!\!$48.7$\!\!$ &               \cellcolor[rgb]{0.9254752851711027, 0.95, 0.9254752851711027} $\!\!$25.4$\!\!$ &  \cellcolor[rgb]{0.9105734767025089, 0.9105734767025089, 0.95} $\!\!$49.1$\!\!$ \\
$\rightarrow$5         &  +subst. &               \cellcolor[rgb]{0.944296577946768, 0.95, 0.944296577946768} $\!\!$22.1$\!\!$ &  \cellcolor[rgb]{0.9124850657108721, 0.9124850657108721, 0.95} $\!\!$47.5$\!\!$ &               \cellcolor[rgb]{0.923764258555133, 0.95, 0.923764258555133} $\!\!$25.7$\!\!$ &  \cellcolor[rgb]{0.9172640382317802, 0.9172640382317802, 0.95} $\!\!$43.5$\!\!$ \\
\cmidrule(r){1-1} \cmidrule(lr){2-2}  \cmidrule{3-6}
2,3,6         &     sys. &               \cellcolor[rgb]{0.9015209125475285, 0.95, 0.9015209125475285} $\!\!$29.6$\!\!$ &  \cellcolor[rgb]{0.9004181600955794, 0.9004181600955794, 0.95} $\!\!$57.6$\!\!$ &               \cellcolor[rgb]{0.870722433460076, 0.95, 0.870722433460076} $\!\!$35.0$\!\!$ &  \cellcolor[rgb]{0.9034050179211469, 0.9034050179211469, 0.95} $\!\!$55.1$\!\!$ \\
$\rightarrow$8         &  +subst. &               \cellcolor[rgb]{0.9066539923954372, 0.95, 0.9066539923954372} $\!\!$28.7$\!\!$ &  \cellcolor[rgb]{0.9090203106332139, 0.9090203106332139, 0.95} $\!\!$50.4$\!\!$ &               \cellcolor[rgb]{0.8747148288973384, 0.95, 0.8747148288973384} $\!\!$34.3$\!\!$ &  \cellcolor[rgb]{0.9092592592592592, 0.9092592592592592, 0.95} $\!\!$50.2$\!\!$ \\
\cmidrule(r){1-1} \cmidrule(lr){2-2}  \cmidrule{3-6}
2,3,6         &     sys. &               \cellcolor[rgb]{0.8610266159695817, 0.95, 0.8610266159695817} $\!\!$36.7$\!\!$ &  \cellcolor[rgb]{0.8977897252090801, 0.8977897252090801, 0.95} $\!\!$59.8$\!\!$ &               \cellcolor[rgb]{0.8399239543726236, 0.95, 0.8399239543726236} $\!\!$40.4$\!\!$ &  \cellcolor[rgb]{0.9112903225806451, 0.9112903225806451, 0.95} $\!\!$48.5$\!\!$ \\
$\rightarrow$7         &  +subst. &               \cellcolor[rgb]{0.8456273764258555, 0.95, 0.8456273764258555} $\!\!$39.4$\!\!$ &  \cellcolor[rgb]{0.9011350059737157, 0.9011350059737157, 0.95} $\!\!$57.0$\!\!$ &               \cellcolor[rgb]{0.850190114068441, 0.95, 0.850190114068441} $\!\!$38.6$\!\!$ &  \cellcolor[rgb]{0.9222819593787336, 0.9222819593787336, 0.95} $\!\!$39.3$\!\!$ \\
\cmidrule(r){1-1} \cmidrule(lr){2-2}  \cmidrule{3-6}
1,2,6         &     sys. &               \cellcolor[rgb]{0.9448669201520912, 0.95, 0.9448669201520912} $\!\!$22.0$\!\!$ &  \cellcolor[rgb]{0.9380525686977299, 0.9380525686977299, 0.95} $\!\!$26.1$\!\!$ &               \cellcolor[rgb]{0.95, 0.95, 0.95} $\!\!$21.1$\!\!$ &  \cellcolor[rgb]{0.95, 0.95, 0.95} $\!\!$16.1$\!\!$ \\
$\rightarrow$9         &  +subst. &               \cellcolor[rgb]{0.9414448669201521, 0.95, 0.9414448669201521} $\!\!$22.6$\!\!$ &  \cellcolor[rgb]{0.9373357228195938, 0.9373357228195938, 0.95} $\!\!$26.7$\!\!$ &               \cellcolor[rgb]{0.9363117870722433, 0.95, 0.9363117870722433} $\!\!$23.5$\!\!$ &  \cellcolor[rgb]{0.9474910394265232, 0.9474910394265232, 0.95} $\!\!$18.2$\!\!$ \\
\cmidrule(r){1-1} \cmidrule(lr){2-2}  \cmidrule{3-6}
7,8 &     sys. &               \cellcolor[rgb]{0.9328897338403042, 0.95, 0.9328897338403042} $\!\!$24.1$\!\!$ &  \cellcolor[rgb]{0.9350657108721624, 0.9350657108721624, 0.95} $\!\!$28.6$\!\!$ &               \cellcolor[rgb]{0.8849809885931559, 0.95, 0.8849809885931559} $\!\!$32.5$\!\!$ &  \cellcolor[rgb]{0.9390083632019115, 0.9390083632019115, 0.95} $\!\!$25.3$\!\!$ \\
$\rightarrow$10 &  +subst. &               \cellcolor[rgb]{0.9283269961977186, 0.95, 0.9283269961977186} $\!\!$24.9$\!\!$ &  \cellcolor[rgb]{0.9361409796893667, 0.9361409796893667, 0.95} $\!\!$27.7$\!\!$ &               \cellcolor[rgb]{0.8712927756653992, 0.95, 0.8712927756653992} $\!\!$34.9$\!\!$ &  \cellcolor[rgb]{0.9342293906810035, 0.9342293906810035, 0.95} $\!\!$29.3$\!\!$ \\
\cmidrule(r){1-1} \cmidrule(lr){2-2}  \cmidrule{3-6}
1         &    prod. &               \cellcolor[rgb]{0.5433460076045626, 0.95, 0.5433460076045626} $\!\!$92.4$\!\!$ &  \cellcolor[rgb]{0.85, 0.85, 0.95} $\!\!$99.8$\!\!$ &              \cellcolor[rgb]{0.5, 0.95, 0.5} $\!\!$100.0$\!\!$ &  \cellcolor[rgb]{0.8699522102747909, 0.8699522102747909, 0.95} $\!\!$83.1$\!\!$ \\
$\rightarrow$3         &  +subst. &               \cellcolor[rgb]{0.5501901140684411, 0.95, 0.5501901140684411} $\!\!$91.2$\!\!$ &  \cellcolor[rgb]{0.85, 0.85, 0.95} $\!\!$99.8$\!\!$ &              \cellcolor[rgb]{0.5, 0.95, 0.5} $\!\!$100.0$\!\!$ &  \cellcolor[rgb]{0.8709080047789725, 0.8709080047789725, 0.95} $\!\!$82.3$\!\!$ \\
\cmidrule(r){1-1} \cmidrule(lr){2-2}  \cmidrule{3-6}
4         &    prod. &               \cellcolor[rgb]{0.7258555133079848, 0.95, 0.7258555133079848} $\!\!$60.4$\!\!$ &  \cellcolor[rgb]{0.8581242532855435, 0.8581242532855435, 0.95} $\!\!$93.0$\!\!$ &               \cellcolor[rgb]{0.5598859315589353, 0.95, 0.5598859315589353} $\!\!$89.5$\!\!$ &  \cellcolor[rgb]{0.860752688172043, 0.860752688172043, 0.95} $\!\!$90.8$\!\!$ \\
$\rightarrow$5         &  +subst. &               \cellcolor[rgb]{0.7064638783269962, 0.95, 0.7064638783269962} $\!\!$63.8$\!\!$ &  \cellcolor[rgb]{0.8546594982078852, 0.8546594982078852, 0.95} $\!\!$95.9$\!\!$ &               \cellcolor[rgb]{0.546768060836502, 0.95, 0.546768060836502} $\!\!$91.8$\!\!$ &  \cellcolor[rgb]{0.8648148148148148, 0.8648148148148148, 0.95} $\!\!$87.4$\!\!$ \\
\cmidrule(r){1-1} \cmidrule(lr){2-2}  \cmidrule{3-6}
9         &    prod. &               \cellcolor[rgb]{0.879277566539924, 0.95, 0.879277566539924} $\!\!$33.5$\!\!$ &  \cellcolor[rgb]{0.9102150537634408, 0.9102150537634408, 0.95} $\!\!$49.4$\!\!$ &               \cellcolor[rgb]{0.8330798479087452, 0.95, 0.8330798479087452} $\!\!$41.6$\!\!$ &  \cellcolor[rgb]{0.9441457586618877, 0.9441457586618877, 0.95} $\!\!$21.0$\!\!$ \\
$\rightarrow$10         &  +subst. &               \cellcolor[rgb]{0.8747148288973384, 0.95, 0.8747148288973384} $\!\!$34.3$\!\!$ &  \cellcolor[rgb]{0.9116487455197132, 0.9116487455197132, 0.95} $\!\!$48.2$\!\!$ &               \cellcolor[rgb]{0.8222433460076045, 0.95, 0.8222433460076045} $\!\!$43.5$\!\!$ &  \cellcolor[rgb]{0.9427120669056153, 0.9427120669056153, 0.95} $\!\!$22.2$\!\!$ \\
\bottomrule
\end{tabular}

\caption{
Experimental results when BART is used as a pre-trained model.
The ``Task'' column exhibits (train$\rightarrow$test) domains corresponding to the skill-tree (Figure~\ref{fig:skill-tree}).
The ``Type'' column shows the targeted compositionality type in each setting; here, ``sys.,'' ``prod,'' and ``subst.'' denote the systematicity, productivity, and substitutivity generalizations, respectively.  
}
\label{table:bart_result}
\end{table}

\section{How to calculate WA}
\label{sec:appendix:how_to_calculate_ws}
As described in Section~\ref{sec:evaluation_method}, we used weighted average accuracy (WA).
This metric quantifies the efficiency (ease) of generalization by assigning a high weight to accuracy in the early training stage.
Specifically, the weights $w_i$ (where $i$ is the number of validation steps) were calculated using the following formulae:

\begin{align}
w_i &= - ai + w_\mathrm{max} \;\;, \\
w_\mathrm{max} &= \alpha w_\mathrm{min} \;\;, \\
w_\mathrm{min} &= \frac{2}{(N+1)(\alpha+1)} \;\;, \\
a &= \frac{(w_{\mathrm{max}}-w_{\mathrm{min}})}{N} \;\;.
\end{align}

\noindent
Here, $N$ is the number of validation steps.
$\alpha$ is a hyperparameter that determines how heavily the accuracy in the early stages is weighted.
We set $\alpha$ to 1000 in all the experiments.
Also, We used the first 100 validation steps to calculate WA, so specify $N=100$ for all experiments.
 WA was calculated with accuracy on the held-out validation datasets in the test domain.

\section{Model variants}
\label{sec:appendix:models}

\subsection{BART}
To confirm the generality of our results obtained with the T5 models, we also conduct the same experiment using BART~\cite{lewis-etal-2020-bart}.
We use two model sizes\footnote{\url{https://huggingface.co/facebook/bart-base}, \url{https://huggingface.co/facebook/bart-large}}: base (140 million parameters) and large (400 million parameters).
Table~\ref{table:bart_result} shows the result using BART.
The same tendency described in Section~\ref{section:results} was observed when BART was used as a language-pre-training model.

\subsection{Training from scratch}
\label{subsec:from_scratch}
We also experimented with the case where we started training from randomly initialized parameters to isolate the effect of the pre-training adopted in T5.
We found that these initialized models failed to learn even primitive operations and in-domain tasks (Table~\ref{table:from_scratch}), at least with the hyperparameter setting\footnote{Unlike in the training domain used for the language pre-trained model, we relax the  training stopping criterion. We stopped training when the accuracy did not increase by \textbf{10} epochs in the validation dataset.} used in this study.
Thus, we did not proceed to their evaluation of compositionality generalization.

\begin{table}[!t]
\centering

\begin{tabular}{lrrr}
\toprule
Task &   base &  large &     x-large \\
\cmidrule(r){1-1} \cmidrule(lr){2-4}
1,2 ($\rightarrow$4) &   72.5 &  100.0 &  100.0 \\
2,3 ($\rightarrow$5)  &   63.3 &   56.1 &   99.9 \\
1,3,6 ($\rightarrow$7, 8)  &   68.6 &   67.9 &   67.5 \\
1,2,6 ($\rightarrow$9)  &   99.9 &   76.0 &   76.8 \\
7,8 ($\rightarrow$10)  &   48.6 &   37.3 &   24.0 \\
1 ($\rightarrow$3)  &   51.1 &   15.1 &   16.3 \\
4 ($\rightarrow$5)    &  100.0 &   99.9 &   99.9 \\
9 ($\rightarrow$10) &   99.7 &   99.9 &   99.7 \\
\bottomrule
\end{tabular}

\caption{
Accuracy on the training held-out dataset when we start training the model from randomly initialized parameters. 
(When training began with parameters pre-trained on language, the accuracy was almost 100\% in all settings.)
Each training task includes the primitive operations required to solve train domain tasks as described in section~\ref{sec:appendix:training_configurations}.
}
\label{table:from_scratch}
\end{table}

\section{Detailed analysis of T5-based models}
\label{appendix:sec:detailed_analysis_of_t5_based_models}

\begin{figure*}[t]
\centering
\includegraphics[width=\linewidth]{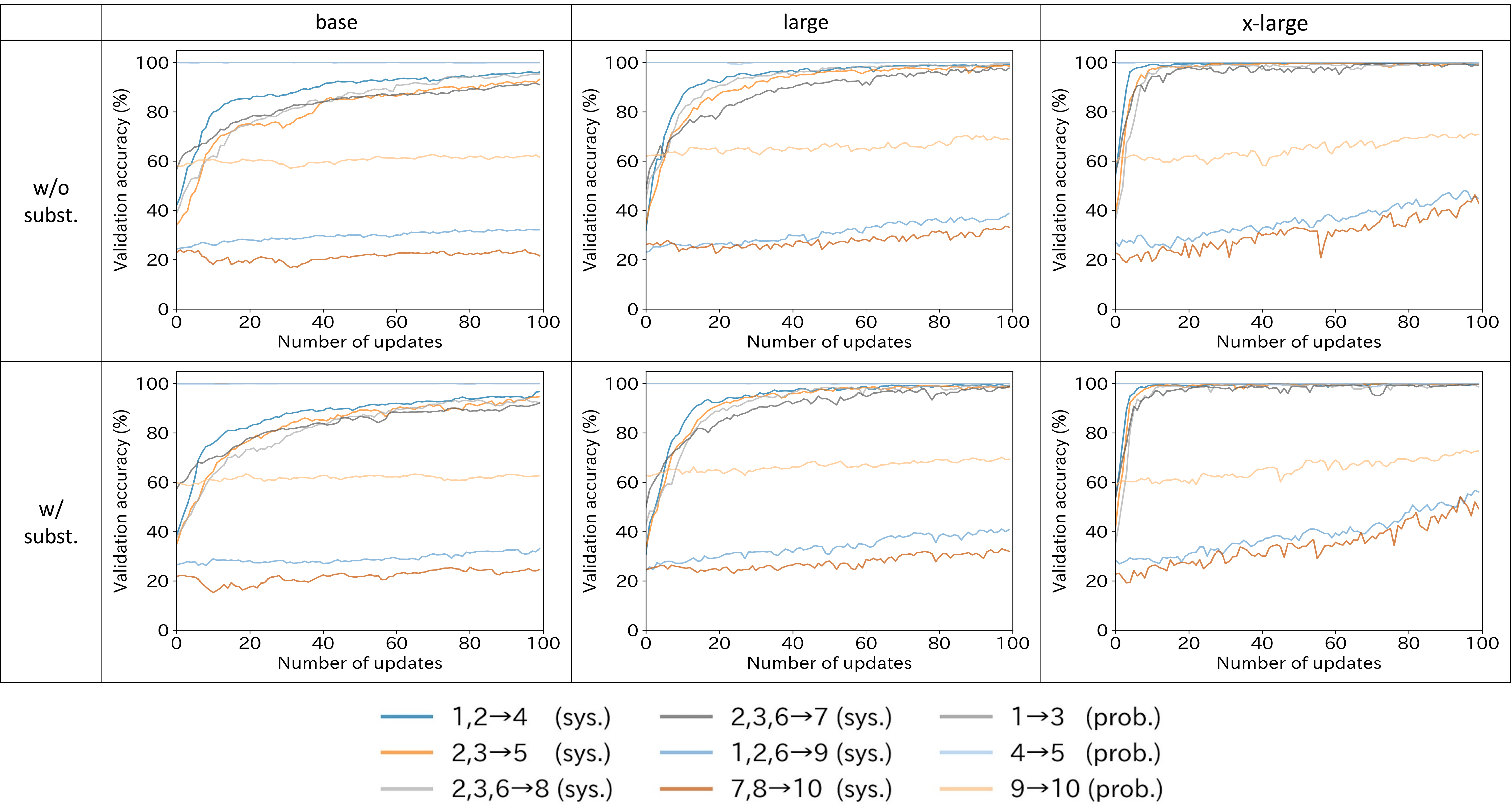}
\caption{Learning curves of generalization test for all tasks.}
\label{fig:learning_curves}  
\end{figure*}

Figure~\ref{fig:learning_curves} shows the learning curves (accuracy on validation datasets) for all tasks demonstrated in Table~\ref{table:compositional_results}. 
The graphs also show that the neural language model struggles to solve tasks that require compositional generalization in the beginning part of the training.

\begin{table*}[!t]
\centering
\begin{tabular}{llrrrrrr}
\toprule
 &  & \multicolumn{2}{c}{base} & \multicolumn{2}{c}{large} & \multicolumn{2}{c}{x-large} \\
Task & Type & ZA &     WS & ZA &     WS & ZA &     WS \\
\cmidrule(r){1-1} \cmidrule(lr){2-2} \cmidrule{3-8}
1,2         &     sys. &               \cellcolor[rgb]{0.7076923076923076, 0.95, 0.7076923076923076} $\!\!$59.2$\!\!$ &   \cellcolor[rgb]{0.8562404870624049, 0.8562404870624049, 0.95} $\!\!$95.9$\!\!$ &               \cellcolor[rgb]{0.6684954751131222, 0.95, 0.6684954751131222} $\!\!$66.9$\!\!$ &   \cellcolor[rgb]{0.852283105022831, 0.852283105022831, 0.95} $\!\!$98.5$\!\!$ &               \cellcolor[rgb]{0.6837669683257919, 0.95, 0.6837669683257919} $\!\!$63.9$\!\!$ &   \cellcolor[rgb]{0.8516742770167427, 0.8516742770167427, 0.95} $\!\!$98.9$\!\!$ \\
$\rightarrow4$         &  +subst. &               \cellcolor[rgb]
{0.7087104072398189, 0.95, 0.7087104072398189} $\!\!$59.0$\!\!$ &   \cellcolor[rgb]{0.8574581430745815, 0.8574581430745815, 0.95} $\!\!$95.1$\!\!$ &               \cellcolor[rgb]{0.6664592760180996, 0.95, 0.6664592760180996} $\!\!$67.3$\!\!$ &   \cellcolor[rgb]{0.8527397260273972, 0.8527397260273972, 0.95} $\!\!$98.2$\!\!$ &               \cellcolor[rgb]{0.692420814479638, 0.95, 0.692420814479638} $\!\!$62.2$\!\!$ &   \cellcolor[rgb]{0.8515220700152206, 0.8515220700152206, 0.95} $\!\!$99.0$\!\!$ \\
\cmidrule(r){1-1} \cmidrule(lr){2-2} \cmidrule{3-8}
$2,3$         &     sys. &               \cellcolor[rgb]{0.8191742081447964, 0.95, 0.8191742081447964} $\!\!$37.3$\!\!$ &   \cellcolor[rgb]{0.8589802130898021, 0.8589802130898021, 0.95} $\!\!$94.1$\!\!$ &               \cellcolor[rgb]{0.6725678733031675, 0.95, 0.6725678733031675} $\!\!$66.1$\!\!$ &   \cellcolor[rgb]{0.8524353120243531, 0.8524353120243531, 0.95} $\!\!$98.4$\!\!$ &               \cellcolor[rgb]{0.5666855203619909, 0.95, 0.5666855203619909} $\!\!$86.9$\!\!$ &   \cellcolor[rgb]{0.8510654490106545, 0.8510654490106545, 0.95} $\!\!$99.3$\!\!$ \\
$\rightarrow5$        &  +subst. &               \cellcolor[rgb]{0.8263009049773755, 0.95, 0.8263009049773755} $\!\!$35.9$\!\!$ &   \cellcolor[rgb]{0.8598934550989346, 0.8598934550989346, 0.95} $\!\!$93.5$\!\!$ &               \cellcolor[rgb]{0.6771493212669684, 0.95, 0.6771493212669684} $\!\!$65.2$\!\!$ &   \cellcolor[rgb]{0.8521308980213089, 0.8521308980213089, 0.95} $\!\!$98.6$\!\!$ &               \cellcolor[rgb]{0.5804298642533936, 0.95, 0.5804298642533936} $\!\!$84.2$\!\!$ &   \cellcolor[rgb]{0.8515220700152206, 0.8515220700152206, 0.95} $\!\!$99.0$\!\!$ \\
\cmidrule(r){1-1} \cmidrule(lr){2-2} \cmidrule{3-8}
2,3,6        &     sys. &               \cellcolor[rgb]{0.705656108597285, 0.95, 0.705656108597285} $\!\!$59.6$\!\!$ &   \cellcolor[rgb]{0.8592846270928463, 0.8592846270928463, 0.95} $\!\!$93.9$\!\!$ &               \cellcolor[rgb]{0.645079185520362, 0.95, 0.645079185520362} $\!\!$71.5$\!\!$ &   \cellcolor[rgb]{0.8530441400304414, 0.8530441400304414, 0.95} $\!\!$98.0$\!\!$ &               \cellcolor[rgb]{0.7417986425339367, 0.95, 0.7417986425339367} $\!\!$52.5$\!\!$ &   \cellcolor[rgb]{0.854261796042618, 0.854261796042618, 0.95} $\!\!$97.2$\!\!$ \\
$\rightarrow$8         &  +subst. &               \cellcolor[rgb]{0.7224547511312218, 0.95, 0.7224547511312218} $\!\!$56.3$\!\!$ &   \cellcolor[rgb]{0.860958904109589, 0.860958904109589, 0.95} $\!\!$92.8$\!\!$ &               \cellcolor[rgb]{0.6674773755656109, 0.95, 0.6674773755656109} $\!\!$67.1$\!\!$ &   \cellcolor[rgb]{0.8530441400304414, 0.8530441400304414, 0.95} $\!\!$98.0$\!\!$ &               \cellcolor[rgb]{0.742816742081448, 0.95, 0.742816742081448} $\!\!$52.3$\!\!$ &   \cellcolor[rgb]{0.8545662100456621, 0.8545662100456621, 0.95} $\!\!$97.0$\!\!$ \\
\cmidrule(r){1-1} \cmidrule(lr){2-2} \cmidrule{3-8}
2,3,6         &     sys. &               \cellcolor[rgb]{0.8344457013574661, 0.95, 0.8344457013574661} $\!\!$34.3$\!\!$ &   \cellcolor[rgb]{0.8636986301369862, 0.8636986301369862, 0.95} $\!\!$91.0$\!\!$ &               \cellcolor[rgb]{0.7733597285067874, 0.95, 0.7733597285067874} $\!\!$46.3$\!\!$ &   \cellcolor[rgb]{0.8597412480974125, 0.8597412480974125, 0.95} $\!\!$93.6$\!\!$ &               \cellcolor[rgb]{0.8461538461538461, 0.95, 0.8461538461538461} $\!\!$32.0$\!\!$ &   \cellcolor[rgb]{0.856544901065449, 0.856544901065449, 0.95} $\!\!$95.7$\!\!$ \\
$\rightarrow$7         &  +subst. &               \cellcolor[rgb]{0.8344457013574661, 0.95, 0.8344457013574661} $\!\!$34.3$\!\!$ &   \cellcolor[rgb]{0.8632420091324201, 0.8632420091324201, 0.95} $\!\!$91.3$\!\!$ &               \cellcolor[rgb]{0.7641968325791855, 0.95, 0.7641968325791855} $\!\!$48.1$\!\!$ &   \cellcolor[rgb]{0.8597412480974125, 0.8597412480974125, 0.95} $\!\!$93.6$\!\!$ &               \cellcolor[rgb]{0.8349547511312216, 0.95, 0.8349547511312216} $\!\!$34.2$\!\!$ &   \cellcolor[rgb]{0.8553272450532724, 0.8553272450532724, 0.95} $\!\!$96.5$\!\!$ \\
\cmidrule(r){1-1} \cmidrule(lr){2-2} \cmidrule{3-8}
1,2,6         &     sys. &               \cellcolor[rgb]{0.9418552036199095, 0.95, 0.9418552036199095} $\!\!$13.2$\!\!$ &   \cellcolor[rgb]{0.95, 0.95, 0.95} $\!\!$34.3$\!\!$ &               \cellcolor[rgb]{0.95, 0.95, 0.95} $\!\!$11.6$\!\!$ &   \cellcolor[rgb]{0.9442161339421613, 0.9442161339421613, 0.95} $\!\!$38.1$\!\!$ &               \cellcolor[rgb]{0.9464366515837104, 0.95, 0.9464366515837104} $\!\!$12.3$\!\!$ &   \cellcolor[rgb]{0.9114916286149163, 0.9114916286149163, 0.95} $\!\!$59.6$\!\!$ \\
$\rightarrow$9         &  +subst. &               \cellcolor[rgb]{0.9444004524886878, 0.95, 0.9444004524886878} $\!\!$12.7$\!\!$ &   \cellcolor[rgb]{0.9490867579908675, 0.9490867579908675, 0.95} $\!\!$34.9$\!\!$ &               \cellcolor[rgb]{0.9428733031674208, 0.95, 0.9428733031674208} $\!\!$13.0$\!\!$ &   \cellcolor[rgb]{0.9428462709284626, 0.9428462709284626, 0.95} $\!\!$39.0$\!\!$ &               \cellcolor[rgb]{0.9459276018099547, 0.95, 0.9459276018099547} $\!\!$12.4$\!\!$ &   \cellcolor[rgb]{0.9127092846270928, 0.9127092846270928, 0.95} $\!\!$58.8$\!\!$ \\
\cmidrule(r){1-1} \cmidrule(lr){2-2} \cmidrule{3-8}
7,8 &     sys. &               \cellcolor[rgb]{0.8756787330316742, 0.95, 0.8756787330316742} $\!\!$26.2$\!\!$ &   \cellcolor[rgb]{0.9092085235920851, 0.9092085235920851, 0.95} $\!\!$61.1$\!\!$ &               \cellcolor[rgb]{0.9072398190045249, 0.95, 0.9072398190045249} $\!\!$20.0$\!\!$ &   \cellcolor[rgb]{0.9082952815829528, 0.9082952815829528, 0.95} $\!\!$61.7$\!\!$ &               \cellcolor[rgb]{0.9006221719457013, 0.95, 0.9006221719457013} $\!\!$21.3$\!\!$ &   \cellcolor[rgb]{0.8996194824961947, 0.8996194824961947, 0.95} $\!\!$67.4$\!\!$ \\
$\rightarrow$10 &  +subst. &               \cellcolor[rgb]{0.8746606334841629, 0.95, 0.8746606334841629} $\!\!$26.4$\!\!$ &   \cellcolor[rgb]{0.908751902587519, 0.908751902587519, 0.95} $\!\!$61.4$\!\!$ &               \cellcolor[rgb]{0.9077488687782805, 0.95, 0.9077488687782805} $\!\!$19.9$\!\!$ &   \cellcolor[rgb]{0.910882800608828, 0.910882800608828, 0.95} $\!\!$60.0$\!\!$ &               \cellcolor[rgb]{0.9046945701357466, 0.95, 0.9046945701357466} $\!\!$20.5$\!\!$ &   \cellcolor[rgb]{0.897945205479452, 0.897945205479452, 0.95} $\!\!$68.5$\!\!$ \\
\cmidrule(r){1-1} \cmidrule(lr){2-2} \cmidrule{3-8}
1         &    prod. &              \cellcolor[rgb]{0.5, 0.95, 0.5} $\!\!$100.0$\!\!$ &  \cellcolor[rgb]{0.85, 0.85, 0.95} $\!\!$100.0$\!\!$ &              \cellcolor[rgb]{0.5, 0.95, 0.5} $\!\!$100.0$\!\!$ &  \cellcolor[rgb]{0.85, 0.85, 0.95} $\!\!$100.0$\!\!$ &              \cellcolor[rgb]{0.5, 0.95, 0.5} $\!\!$100.0$\!\!$ &  \cellcolor[rgb]{0.85, 0.85, 0.95} $\!\!$100.0$\!\!$ \\
$\rightarrow$3         &  +subst. &              \cellcolor[rgb]{0.5, 0.95, 0.5} $\!\!$100.0$\!\!$ &  \cellcolor[rgb]{0.85, 0.85, 0.95} $\!\!$100.0$\!\!$ &              \cellcolor[rgb]{0.5, 0.95, 0.5} $\!\!$100.0$\!\!$ &  \cellcolor[rgb]{0.85, 0.85, 0.95} $\!\!$100.0$\!\!$ &               \cellcolor[rgb]{0.5005090497737557, 0.95, 0.5005090497737557} $\!\!$99.9$\!\!$ &  \cellcolor[rgb]{0.85, 0.85, 0.95} $\!\!$100.0$\!\!$ \\
\cmidrule(r){1-1} \cmidrule(lr){2-2} \cmidrule{3-8}
4         &    prod. &              \cellcolor[rgb]{0.5, 0.95, 0.5} $\!\!$100.0$\!\!$ &  \cellcolor[rgb]{0.85, 0.85, 0.95} $\!\!$100.0$\!\!$ &               \cellcolor[rgb]{0.5005090497737557, 0.95, 0.5005090497737557} $\!\!$99.9$\!\!$ &  \cellcolor[rgb]{0.85, 0.85, 0.95} $\!\!$100.0$\!\!$ &              \cellcolor[rgb]{0.5, 0.95, 0.5} $\!\!$100.0$\!\!$ &  \cellcolor[rgb]{0.85, 0.85, 0.95} $\!\!$100.0$\!\!$ \\
$\rightarrow$5         &  +subst. &              \cellcolor[rgb]{0.5, 0.95, 0.5} $\!\!$100.0$\!\!$ &  \cellcolor[rgb]{0.85, 0.85, 0.95} $\!\!$100.0$\!\!$ &              \cellcolor[rgb]{0.5, 0.95, 0.5} $\!\!$100.0$\!\!$ &  \cellcolor[rgb]{0.85, 0.85, 0.95} $\!\!$100.0$\!\!$ &              \cellcolor[rgb]{0.5, 0.95, 0.5} $\!\!$100.0$\!\!$ &  \cellcolor[rgb]{0.85, 0.85, 0.95} $\!\!$100.0$\!\!$ \\
\cmidrule(r){1-1} \cmidrule(lr){2-2} \cmidrule{3-8}
9         &    prod. &               \cellcolor[rgb]{0.6547511312217196, 0.95, 0.6547511312217196} $\!\!$69.6$\!\!$ &   \cellcolor[rgb]{0.8679604261796042, 0.8679604261796042, 0.95} $\!\!$88.2$\!\!$ &               \cellcolor[rgb]{0.645079185520362, 0.95, 0.645079185520362} $\!\!$71.5$\!\!$ &   \cellcolor[rgb]{0.865372907153729, 0.865372907153729, 0.95} $\!\!$89.9$\!\!$ &               \cellcolor[rgb]{0.6257352941176471, 0.95, 0.6257352941176471} $\!\!$75.3$\!\!$ &   \cellcolor[rgb]{0.8621765601217656, 0.8621765601217656, 0.95} $\!\!$92.0$\!\!$ \\
$\rightarrow$10         &  +subst. &               \cellcolor[rgb]{0.665950226244344, 0.95, 0.665950226244344} $\!\!$67.4$\!\!$ &   \cellcolor[rgb]{0.8670471841704718, 0.8670471841704718, 0.95} $\!\!$88.8$\!\!$ &               \cellcolor[rgb]{0.6425339366515838, 0.95, 0.6425339366515838} $\!\!$72.0$\!\!$ &   \cellcolor[rgb]{0.8658295281582953, 0.8658295281582953, 0.95} $\!\!$89.6$\!\!$ &               \cellcolor[rgb]{0.6262443438914027, 0.95, 0.6262443438914027} $\!\!$75.2$\!\!$ &   \cellcolor[rgb]{0.8621765601217656, 0.8621765601217656, 0.95} $\!\!$92.0$\!\!$ \\
\bottomrule
\end{tabular}

\caption{
Experiments result from string operation ablation.
The ``Task'' column exhibits (train$\rightarrow$test) domains corresponding to the skill-tree (Figure~\ref{fig:skill-tree}).
The ``Type'' column shows the targeted compositionality type in each setting; here, ``sys.,'' ``prod,'' and ``subst.'' denote the systematicity, productivity, and substitutivity generalizations, respectively.  
}
\label{table:str_ablation}
\end{table*}

\begin{table*}[!t]
\centering
\begin{tabular}{llllllll}
\toprule
 &  & \multicolumn{2}{c}{base} & \multicolumn{2}{c}{large} & \multicolumn{2}{c}{x-large} \\
Task & Type & ZA &     WS & ZA &     WS & ZA &     WS \\
\cmidrule(r){1-1} \cmidrule(lr){2-2} \cmidrule{3-8}
1,2 &     sys.  &               \cellcolor[rgb]{0.8123, 0.95, 0.8123} $\!\!$30.6$\!\!$ &   \cellcolor[rgb]{0.8687187187187186, 0.8687187187187186, 0.95} $\!\!$81.3$\!\!$ &               \cellcolor[rgb]{0.79745, 0.95, 0.79745} $\!\!$33.9$\!\!$ &   \cellcolor[rgb]{0.8645145145145144, 0.8645145145145144, 0.95} $\!\!$85.5$\!\!$ &               \cellcolor[rgb]{0.7898, 0.95, 0.7898} $\!\!$35.6$\!\!$ &  \cellcolor[rgb]{0.8551051051051051, 0.8551051051051051, 0.95} $\!\!$94.9$\!\!$ \\
$\rightarrow$4         &  +subst. &               \cellcolor[rgb]{0.81365, 0.95, 0.81365} $\!\!$30.3$\!\!$ &   \cellcolor[rgb]{0.872122122122122, 0.872122122122122, 0.95} $\!\!$77.9$\!\!$ &               \cellcolor[rgb]{0.8019499999999999, 0.95, 0.8019499999999999} $\!\!$32.9$\!\!$ &   \cellcolor[rgb]{0.8645145145145144, 0.8645145145145144, 0.95} $\!\!$85.5$\!\!$ &               \cellcolor[rgb]{0.78665, 0.95, 0.78665} $\!\!$36.3$\!\!$ &  \cellcolor[rgb]{0.8545045045045044, 0.8545045045045044, 0.95} $\!\!$95.5$\!\!$ \\
\cmidrule(r){1-1} \cmidrule(lr){2-2} \cmidrule{3-8}
2,3         &     sys. &               \cellcolor[rgb]{0.8321, 0.95, 0.8321} $\!\!$26.2$\!\!$ &   \cellcolor[rgb]{0.867917917917918, 0.867917917917918, 0.95} $\!\!$82.1$\!\!$ &               \cellcolor[rgb]{0.78755, 0.95, 0.78755} $\!\!$36.1$\!\!$ &   \cellcolor[rgb]{0.8606106106106106, 0.8606106106106106, 0.95} $\!\!$89.4$\!\!$ &               \cellcolor[rgb]{0.7983499999999999, 0.95, 0.7983499999999999} $\!\!$33.7$\!\!$ &  \cellcolor[rgb]{0.8536036036036035, 0.8536036036036035, 0.95} $\!\!$96.4$\!\!$ \\
$\rightarrow$5         &  +subst. &               \cellcolor[rgb]{0.8357, 0.95, 0.8357} $\!\!$25.4$\!\!$ &   \cellcolor[rgb]{0.8687187187187186, 0.8687187187187186, 0.95} $\!\!$81.3$\!\!$ &               \cellcolor[rgb]{0.78755, 0.95, 0.78755} $\!\!$36.1$\!\!$ &   \cellcolor[rgb]{0.8619119119119119, 0.8619119119119119, 0.95} $\!\!$88.1$\!\!$ &               \cellcolor[rgb]{0.8019499999999999, 0.95, 0.8019499999999999} $\!\!$32.9$\!\!$ &  \cellcolor[rgb]{0.8536036036036035, 0.8536036036036035, 0.95} $\!\!$96.4$\!\!$ \\
\cmidrule(r){1-1} \cmidrule(lr){2-2} \cmidrule{3-8}
2,3,6         &     sys. &                \cellcolor[rgb]{0.95, 0.95, 0.95} $\!\!$0.0$\!\!$ &   \cellcolor[rgb]{0.9224724724724724, 0.9224724724724724, 0.95} $\!\!$27.6$\!\!$ &                \cellcolor[rgb]{0.95, 0.95, 0.95} $\!\!$0.0$\!\!$ &   \cellcolor[rgb]{0.9025525525525525, 0.9025525525525525, 0.95} $\!\!$47.5$\!\!$ &                \cellcolor[rgb]{0.95, 0.95, 0.95} $\!\!$0.0$\!\!$ &  \cellcolor[rgb]{0.8787287287287286, 0.8787287287287286, 0.95} $\!\!$71.3$\!\!$ \\
$\rightarrow$8         &  +subst. &                \cellcolor[rgb]{0.95, 0.95, 0.95} $\!\!$0.0$\!\!$ &   \cellcolor[rgb]{0.924074074074074, 0.924074074074074, 0.95} $\!\!$26.0$\!\!$ &                \cellcolor[rgb]{0.95, 0.95, 0.95} $\!\!$0.0$\!\!$ &   \cellcolor[rgb]{0.9027527527527527, 0.9027527527527527, 0.95} $\!\!$47.3$\!\!$ &                \cellcolor[rgb]{0.95, 0.95, 0.95} $\!\!$0.0$\!\!$ &  \cellcolor[rgb]{0.877127127127127, 0.877127127127127, 0.95} $\!\!$72.9$\!\!$ \\
\cmidrule(r){1-1} \cmidrule(lr){2-2} \cmidrule{3-8}
2,3,6         &     sys. &                \cellcolor[rgb]{0.95, 0.95, 0.95} $\!\!$0.0$\!\!$ &   \cellcolor[rgb]{0.9069569569569569, 0.9069569569569569, 0.95} $\!\!$43.1$\!\!$ &                \cellcolor[rgb]{0.95, 0.95, 0.95} $\!\!$0.0$\!\!$ &   \cellcolor[rgb]{0.891041041041041, 0.891041041041041, 0.95} $\!\!$59.0$\!\!$ &                \cellcolor[rgb]{0.95, 0.95, 0.95} $\!\!$0.0$\!\!$ &  \cellcolor[rgb]{0.875025025025025, 0.875025025025025, 0.95} $\!\!$75.0$\!\!$ \\
$\rightarrow$7         &  +subst. &                \cellcolor[rgb]{0.95, 0.95, 0.95} $\!\!$0.0$\!\!$ &   \cellcolor[rgb]{0.91006006006006, 0.91006006006006, 0.95} $\!\!$40.0$\!\!$ &                \cellcolor[rgb]{0.95, 0.95, 0.95} $\!\!$0.0$\!\!$ &   \cellcolor[rgb]{0.8934434434434434, 0.8934434434434434, 0.95} $\!\!$56.6$\!\!$ &                \cellcolor[rgb]{0.95, 0.95, 0.95} $\!\!$0.0$\!\!$ &  \cellcolor[rgb]{0.8767267267267267, 0.8767267267267267, 0.95} $\!\!$73.3$\!\!$ \\
\cmidrule(r){1-1} \cmidrule(lr){2-2} \cmidrule{3-8}
1,2,6         &     sys. &                \cellcolor[rgb]{0.95, 0.95, 0.95} $\!\!$0.0$\!\!$ &    \cellcolor[rgb]{0.9498998998998999, 0.9498998998998999, 0.95} $\!\!$0.2$\!\!$ &                \cellcolor[rgb]{0.95, 0.95, 0.95} $\!\!$0.0$\!\!$ &    \cellcolor[rgb]{0.9475975975975975, 0.9475975975975975, 0.95} $\!\!$2.5$\!\!$ &                \cellcolor[rgb]{0.95, 0.95, 0.95} $\!\!$0.0$\!\!$ &  \cellcolor[rgb]{0.9357857857857858, 0.9357857857857858, 0.95} $\!\!$14.3$\!\!$ \\
$\rightarrow$9         &  +subst. &                \cellcolor[rgb]{0.95, 0.95, 0.95} $\!\!$0.0$\!\!$ &    \cellcolor[rgb]{0.95, 0.95, 0.95} $\!\!$0.1$\!\!$ &                \cellcolor[rgb]{0.95, 0.95, 0.95} $\!\!$0.0$\!\!$ &    \cellcolor[rgb]{0.9473973973973974, 0.9473973973973974, 0.95} $\!\!$2.7$\!\!$ &                \cellcolor[rgb]{0.95, 0.95, 0.95} $\!\!$0.0$\!\!$ &  \cellcolor[rgb]{0.9336836836836836, 0.9336836836836836, 0.95} $\!\!$16.4$\!\!$ \\
\cmidrule(r){1-1} \cmidrule(lr){2-2} \cmidrule{3-8}
7,8 &     sys. &                \cellcolor[rgb]{0.95, 0.95, 0.95} $\!\!$0.0$\!\!$ &   \cellcolor[rgb]{0.9330830830830831, 0.9330830830830831, 0.95} $\!\!$17.0$\!\!$ &                \cellcolor[rgb]{0.95, 0.95, 0.95} $\!\!$0.0$\!\!$ &   \cellcolor[rgb]{0.9168668668668668, 0.9168668668668668, 0.95} $\!\!$33.2$\!\!$ &                \cellcolor[rgb]{0.95, 0.95, 0.95} $\!\!$0.0$\!\!$ &  \cellcolor[rgb]{0.8949449449449449, 0.8949449449449449, 0.95} $\!\!$55.1$\!\!$ \\
$\rightarrow$10 &  +subst. &                \cellcolor[rgb]{0.95, 0.95, 0.95} $\!\!$0.0$\!\!$ &   \cellcolor[rgb]{0.9326826826826826, 0.9326826826826826, 0.95} $\!\!$17.4$\!\!$ &                \cellcolor[rgb]{0.95, 0.95, 0.95} $\!\!$0.0$\!\!$ &   \cellcolor[rgb]{0.9174674674674674, 0.9174674674674674, 0.95} $\!\!$32.6$\!\!$ &                \cellcolor[rgb]{0.95, 0.95, 0.95} $\!\!$0.0$\!\!$ &  \cellcolor[rgb]{0.8947447447447447, 0.8947447447447447, 0.95} $\!\!$55.3$\!\!$ \\
\cmidrule(r){1-1} \cmidrule(lr){2-2} \cmidrule{3-8}
1         &    prod. &              \cellcolor[rgb]{0.5, 0.95, 0.5} $\!\!$100.0$\!\!$ &  \cellcolor[rgb]{0.85, 0.85, 0.95} $\!\!$100.0$\!\!$ &              \cellcolor[rgb]{0.5, 0.95, 0.5} $\!\!$100.0$\!\!$ &  \cellcolor[rgb]{0.85, 0.85, 0.95} $\!\!$100.0$\!\!$ &               \cellcolor[rgb]{0.518, 0.95, 0.518} $\!\!$96.0$\!\!$ &  \cellcolor[rgb]{0.8501001001001001, 0.8501001001001001, 0.95} $\!\!$99.9$\!\!$ \\
$\rightarrow$3         &  +subst. &              \cellcolor[rgb]{0.5, 0.95, 0.5} $\!\!$100.0$\!\!$ &  \cellcolor[rgb]{0.85, 0.85, 0.95} $\!\!$100.0$\!\!$ &              \cellcolor[rgb]{0.5, 0.95, 0.5} $\!\!$100.0$\!\!$ &  \cellcolor[rgb]{0.85, 0.85, 0.95} $\!\!$100.0$\!\!$ &               \cellcolor[rgb]{0.5189, 0.95, 0.5189} $\!\!$95.8$\!\!$ &  \cellcolor[rgb]{0.8501001001001001, 0.8501001001001001, 0.95} $\!\!$99.9$\!\!$ \\
\cmidrule(r){1-1} \cmidrule(lr){2-2} \cmidrule{3-8}
4         &    prod. &              \cellcolor[rgb]{0.5, 0.95, 0.5} $\!\!$100.0$\!\!$ &  \cellcolor[rgb]{0.85, 0.85, 0.95} $\!\!$100.0$\!\!$ &              \cellcolor[rgb]{0.5, 0.95, 0.5} $\!\!$100.0$\!\!$ &   \cellcolor[rgb]{0.8501001001001001, 0.8501001001001001, 0.95} $\!\!$99.9$\!\!$ &              \cellcolor[rgb]{0.5, 0.95, 0.5} $\!\!$100.0$\!\!$ &  \cellcolor[rgb]{0.8501001001001001, 0.8501001001001001, 0.95} $\!\!$99.9$\!\!$ \\
$\rightarrow$5         &  +subst. &              \cellcolor[rgb]{0.5, 0.95, 0.5} $\!\!$100.0$\!\!$ &  \cellcolor[rgb]{0.85, 0.85, 0.95} $\!\!$100.0$\!\!$ &              \cellcolor[rgb]{0.5, 0.95, 0.5} $\!\!$100.0$\!\!$ &   \cellcolor[rgb]{0.8501001001001001, 0.8501001001001001, 0.95} $\!\!$99.9$\!\!$ &              \cellcolor[rgb]{0.5, 0.95, 0.5} $\!\!$100.0$\!\!$ &  \cellcolor[rgb]{0.8501001001001001, 0.8501001001001001, 0.95} $\!\!$99.9$\!\!$ \\
\cmidrule(r){1-1} \cmidrule(lr){2-2} \cmidrule{3-8}
9         &    prod. &                \cellcolor[rgb]{0.95, 0.95, 0.95} $\!\!$0.0$\!\!$ &   \cellcolor[rgb]{0.8992492492492492, 0.8992492492492492, 0.95} $\!\!$50.8$\!\!$ &                \cellcolor[rgb]{0.95, 0.95, 0.95} $\!\!$0.0$\!\!$ &   \cellcolor[rgb]{0.8778278278278278, 0.8778278278278278, 0.95} $\!\!$72.2$\!\!$ &                \cellcolor[rgb]{0.95, 0.95, 0.95} $\!\!$0.0$\!\!$ &  \cellcolor[rgb]{0.8691191191191191, 0.8691191191191191, 0.95} $\!\!$80.9$\!\!$ \\
$\rightarrow$10         &  +subst. &                \cellcolor[rgb]{0.95, 0.95, 0.95} $\!\!$0.0$\!\!$ &   \cellcolor[rgb]{0.8982482482482482, 0.8982482482482482, 0.95} $\!\!$51.8$\!\!$ &                \cellcolor[rgb]{0.95, 0.95, 0.95} $\!\!$0.0$\!\!$ &   \cellcolor[rgb]{0.8805305305305305, 0.8805305305305305, 0.95} $\!\!$69.5$\!\!$ &                \cellcolor[rgb]{0.95, 0.95, 0.95} $\!\!$0.0$\!\!$ &  \cellcolor[rgb]{0.8707207207207207, 0.8707207207207207, 0.95} $\!\!$79.3$\!\!$ \\
\bottomrule
\end{tabular}

\caption{
Experiments result from scratchpad ablation.
The ``Task'' column exhibits (train$\rightarrow$test) domains corresponding to the skill-tree (Figure~\ref{fig:skill-tree}).
The ``Type'' column shows the targeted compositionality type in each setting; here, ``sys.,'' ``prod,'' and ``subst.'' denote the systematicity, productivity, and substitutivity generalizations, respectively. 
}
\label{table:scrtachpad_ablation}
\end{table*}

\subsection{String operations ablation for all tasks}
\label{subsec:str_ablation_for_all_task}
Table~\ref{table:str_ablation} shows the experimental results of the ablation study using string operations for all tasks.

\subsection{Scratch pad evaluation for all tasks}
\label{subsec:scratch_pad_ablation_for_all_task}
Table~\ref{table:scrtachpad_ablation} shows the experimental results of the ablation study using scratchpad for all tasks.

\section{Dataset details}
\label{sec:appendix:dataset_details}

\begin{table*}[t]
\centering
\begin{tabular}{llllllll}
\toprule
\multicolumn{2}{l}{Domains}         & \multicolumn{3}{l}{Required premitives} & \multicolumn{3}{l}{Targeted composition}                                                         \\ 
\cmidrule(){1-2} \cmidrule(lr){3-5}  \cmidrule(lr){6-8}
Task & Example                           &  Assign.     &  Arith.    &  Ref.  & Sys.                                                     & Prod. & Subst. \\ \hline
1 & \texttt{A=1, B=2, B=?}            & \checkmark      &                      &               &                                                                  &              &                \\
2 & \texttt{A=1+2, A=?}               & \checkmark    & \checkmark           &               &                                                                  &              &                \\
3 & \texttt{A=1, B=2, C=3, C=?}       & \checkmark      &                      &               &                                                                  & \checkmark   &                \\
3' & \texttt{α=1, β=2, γ=3, γ=?}       & \checkmark      &                      &               &                                                                  & \checkmark   & \checkmark     \\
4 & \texttt{A=1+2, B=2+3, B=?}        & \checkmark      & \checkmark           &               & \checkmark   &              &                \\
4' & \texttt{α=1+2, β=2+3, β=?}        & \checkmark      & \checkmark           &               & \begin{tabular}[c]{@{}l@{}}\checkmark \end{tabular}     &              & \checkmark     \\
5 & \texttt{A=1+2, B=2+3, C=3+4, C=?} & \checkmark      & \checkmark           &               & \begin{tabular}[c]{@{}l@{}}\checkmark \end{tabular}     & \checkmark   &                \\
5' & \texttt{α=1+2, β=2+3, γ=3+4, γ=?} & \checkmark      & \checkmark           &               & \begin{tabular}[c]{@{}l@{}}\checkmark \end{tabular}     & \checkmark   & \checkmark     \\
6 & \texttt{A=1, B=A, B=?}            & \checkmark    &                      & \checkmark    &                                                                  &              &                \\
7 & \texttt{A=1, B=2, C=B+3, C=?}     & \checkmark      & \checkmark           & \checkmark    & \begin{tabular}[c]{@{}l@{}}\checkmark \end{tabular} &              &                \\
7' & \texttt{α=1, β=2, γ=β+3, γ=?}     & \checkmark      & \checkmark           & \checkmark    & \begin{tabular}[c]{@{}l@{}}\checkmark \end{tabular} &              & \checkmark     \\
8 & \texttt{A=1+2, B=2+3, C=B, C=?}   & \checkmark      & \checkmark           & \checkmark    & \begin{tabular}[c]{@{}l@{}}\checkmark \end{tabular} &              &                \\
8' & \texttt{α=1+2, β=2+3, γ=β, γ=?}   & \checkmark      & \checkmark           & \checkmark    & \begin{tabular}[c]{@{}l@{}}\checkmark \end{tabular} &              & \checkmark     \\
9 & \texttt{A=1+2, B=A+3, B=?}        & \checkmark      & \checkmark           & \checkmark    &  \checkmark  &              &                \\
9' & \texttt{α=1+2, β=α+3, β=?}        & \checkmark      & \checkmark           & \checkmark    & \checkmark &              &  \checkmark  \\
10 & \texttt{A=1+2, B=A+3, C=B+4, C=?} & \checkmark      & \checkmark           & \checkmark    &  \checkmark & \checkmark   &                \\
10' & \texttt{α=1+2, β=α+3, γ=β+4, γ=?} & \checkmark      & \checkmark           & \checkmark    & \checkmark & \checkmark   & \checkmark     \\
\bottomrule
\end{tabular}

\caption{
Dataset configurations. 
The ``Task'' column exhibits (train$\rightarrow$test) domains corresponding to the skill-tree (Figure~\ref{fig:skill-tree}).
``Assign.'' ``Arith.'' and ``Ref.'' in ``Required primitives'' column mean primitive operations name, ``assignment'' ``arithmetic operations'' and ``reference'' (see subsection ~\ref{subsec:dataset_configurations_and_evaluation_methods})
The ``Target composition'' column shows the targeted compositionality type in each setting; here, ``Sys.,'' ``Prod,'' and ``Subst.'' denote the systematicity, productivity, and substitutivity generalizations, respectively.  
}
\label{table:data-configurations}
\end{table*}

\begin{table*}[!t]
\centering
\begin{tabular}{lp{5cm}p{3cm}}
\toprule
 & Description & Example \\
\midrule
\texttt{join} & String concatenation. & \texttt{12 + 34 = 1234} \\
 \cmidrule(lr){2-2} \cmidrule{3-3}
\texttt{reverseJoin} & String concatenation + Reverse & \texttt{123 \textasciicircum\ 78 = 87321} \\
\cmidrule(r){1-1} \cmidrule(lr){2-2} \cmidrule{3-3}
\texttt{strSub} & Deletion of duplicate characters. Return 0 if there are no duplicates.  & \texttt{7873 - 73 = 87} \\
\cmidrule(r){1-1} \cmidrule(lr){2-2} \cmidrule{3-3}
\texttt{stackJoin} & Select one character from the left side of each string alternately and return the combined string. & \texttt{12 * 34 = 1324} \\
\bottomrule
\end{tabular}

\caption{
Detail of string operation.
}
\label{table:str_operations_configurations}
\end{table*}

Table~\ref{table:data-configurations} summarizes the characteristics of each task.
\subsection{String operations}
\label{appendix:subsec:straing_operations}
Table~\ref{table:str_operations_configurations} shows the details of the string operations.

\subsection{Scratch pad formulation}
\label{appendix:subsec:scratch_pad_formulation}

For the additional study using scratchpad in section~\ref{section:results}, we generate problems with ing the intermediate steps.
Scratchpad is in the form of straightforward one-by-one calculations of only those calculations necessary to find the target variable.
The following is an example of scratchpad reasoning:
\graybox{
\emph{Question:} \texttt{A=1+2, B=A+3, B=?} \\
\emph{Answer:} \texttt{A=1+2;A=3;B=A+3;B=3+3;B=6}
}

\end{document}